\documentclass{article} % For LaTeX2e
\usepackage{iclr2026_conference,times}
\usepackage[utf8]{inputenc}
\usepackage[T1]{fontenc}
\usepackage{amsmath, amssymb, amsthm}
\usepackage{tabularx}
\usepackage{amsfonts}
\usepackage{graphicx}
\usepackage{booktabs}
\usepackage{caption}
\usepackage{subcaption}
\usepackage[colorlinks=true]{hyperref}
\usepackage{url}
\usepackage{soul}
\usepackage{natbib} % or biblatex or whatever you use

\soulregister\citet{7}
\soulregister\citep{7}
\soulregister\cite{7}
\soulregister\cref{7}
\soulregister\ref{7}

\usepackage{nicefrac}
\usepackage{microtype}
\usepackage{enumitem}
\usepackage{algorithm}
\usepackage{algpseudocode}
\usepackage{adjustbox}
\usepackage{siunitx}
\usepackage{setspace}
\usepackage{xspace}
\usepackage{wrapfig}
\usepackage[acronym,nowarn]{glossaries}
\glsdisablehyper
\makeglossaries

\usepackage[dvipsnames,table]{xcolor}

\usepackage{hyperref}
\hypersetup{colorlinks,citecolor=blue!50!green,linkcolor=red!80!black,urlcolor=blue}
\usepackage[capitalize,nameinlink]{cleveref}
\crefname{section}{\S}{\S\S}
\Crefname{section}{\S}{\S\S}
\creflabelformat{equation}{#2\textup{#1}#3}

\newcommand{\da}{\textsc{da}\xspace}

\newcommand{\mathbold}[1]{{\boldsymbol{\mathbf{#1}}}}

\newcommand{\g}{\,|\,}

\newcommand{\nestedmathbold}[1]{{\mathbold{#1}}}

\newcommand{\mbx}{\nestedmathbold{x}}
\newcommand{\mby}{\nestedmathbold{y}}

\newcommand{\mbX}{\nestedmathbold{X}}
\newcommand{\mbY}{\nestedmathbold{Y}}

\newcommand{\mbgamma}{\nestedmathbold{\gamma}}

\newcommand{\mbphi}{\nestedmathbold{\phi}}

\newcommand{\mbtheta}{\nestedmathbold{\theta}}

\newcommand{\cD}{\mathcal{D}}

\newcommand{\E}{\mathbb{E}}

\newacronym{KL}{\textsc{kl}}{Kullback-Leibler}

\newacronym{MCMC}{\textsc{mcmc}}{Markov chain Monte Carlo}
\newcommand{\MCMC}{\gls{MCMC}\xspace}

\newacronym[firstplural=Bayesian neural networks]{BNN}{\textsc{bnn}}{Bayesian neural network}

\newacronym{ECE}{\textsc{ece}}{Expected Calibration Error}
\newacronym{OOD}{\textsc{ood}}{out-of-distribution}
\newacronym{ACC}{\textsc{acc}}{Accuracy}
\newacronym{ELBO}{\textsc{elbo}}{Evidence Lower BOund}
\newacronym{MC}{\textsc{mc}}{Monte Carlo}

\newcommand{\ECE}{\gls{ECE}\xspace}

\newcommand{\ELBO}{\gls{ELBO}\xspace}
\newcommand{\MC}{\gls{MC}\xspace}

\newcommand{\cifar}{\textsc{cifar}\textsc{\footnotesize{10}}\xspace}

\newcommand{\imagenet}{\textsc{ImageNet}\xspace}
\newcommand{\imagenetc}{\textsc{ImageNet-C}\xspace}

\newtheorem{theorem}{Theorem}[section]
\newtheorem{proposition}[theorem]{Proposition}

\newtheorem{corollary}[theorem]{Corollary}
\theoremstyle{definition}
\newtheorem{definition}[theorem]{Definition}
\theoremstyle{remark}
\newtheorem{remark}[theorem]{Remark}

\usepackage{tcolorbox}

\definecolor{mylightgray}{gray}{0.95}
\definecolor{boxcolor}{HTML}{faf9f5}
\newtcolorbox{mybox}{colback=mylightgray,colframe=mylightgray,top=0.8pt,bottom=0.8pt,right=1.8pt,left=1.8pt}

\newcommand{\R}{\mathbb{R}}
\newcommand{\N}{\mathcal{N}}
\newcommand{\KL}{\text{KL}}

\newcommand{\ourmethod}{\textsc{OPTIMA}\xspace}

\title{Optimizing Data Augmentation \\ through Bayesian Model Selection}

\iclrfinalcopy 
\author{%
  Madi Matymov\thanks{Corresponding author: \texttt{madi.matymov@kaust.edu.sa}} \\
  KAUST \\ Saudi Arabia 
  \And
  Ba-Hien Tran \\
  Huawei Paris Research Center \\ France 
  \And
  Michael Kampffmeyer \\
  UiT The Arctic University of Norway \\ Norwegian Computing Center, Norway
  \And
  Markus Heinonen \\
  Aalto University\\ Finland 
  \And
  Maurizio Filippone \\
  KAUST\\ Saudi Arabia 
}

\begin{document}

\addtocontents{toc}{\protect\setcounter{tocdepth}{0}}

\maketitle

\begin{abstract}
Data Augmentation (\da) has become an essential tool to improve robustness and generalization of modern machine learning. 
However, when deciding on \da strategies it is critical to choose parameters carefully, and this can be a daunting task which is traditionally left to trial-and-error or expensive optimization based on validation performance.
In this paper, we counter these limitations by proposing a novel framework for optimizing \da. 
In particular, we take a probabilistic view of \da, which leads to the interpretation of augmentation parameters as model (hyper)-parameters, and the optimization of the marginal likelihood with respect to these parameters as a Bayesian model selection problem. 
Due to its intractability, we derive a tractable \ELBO, which allows us to optimize augmentation parameters jointly with model parameters. We provide extensive theoretical results on variational approximation quality, generalization guarantees, invariance properties, and connections to empirical Bayes.
Through experiments on computer vision and NLP tasks, we show that our approach improves calibration and yields robust performance over fixed or no augmentation. 
Our work provides a rigorous foundation for optimizing \da through Bayesian principles with significant potential for robust machine learning.
\end{abstract}

\vspace{-1.8ex}

%------------------------------------------------------------------------------%
\section{Introduction}
\label{sec:intro}

\vspace{-1.5ex}

Data Augmentation (\da) \citep{van2001art} is an essential element behind the success of modern machine learning \cite[see, e.g.,][and references therein]{shorten2019survey}. 
In supervised learning, \da amounts to creating copies of the data in the training set, and perturbing these with sensible transformations that preserve label information.   
The success of \da is connected with the current trend of employing over-parameterized models based on neural networks, which require large amounts of data to be trained effectively \citep{alabdulmohsin2022revisiting}. 
It has been shown that \da has strong connections with regularization \citep{zhang2017understanding,dao2019kernel}, and it can provide a better estimation of the expected risk \citep{shao2022theory,chen2020group,lyle2020benefits,deng2022strong}. 
Therefore, it is expected for \da to enhance generalization.

%%\vspace{-0.5ex}

For a given problem, once transformation for \da are chosen, it is then necessary to decide on their parameters. 
For example, in image classification, if we choose to apply transformations in the form of rotations, what range of angles should we choose?
Careful choices of \da parameters are important to obtain performance improvements. 
For example, in the case of rotations applied to the popular \textsc{mnist} dataset, large rotation angles can turn a '9' into a '6', negatively impacting training. 
In the literature, \da parameters are often suggested after some trial-and-error. 
Direct optimization of \da parameters could also be approached via grid-search or Bayesian optimization by recording performance on a validation set, but this is very costly due to the need to perform a large number of training runs.

\begin{figure}[t]
    \centering
    % Adjust the width or path to match your setup
    \includegraphics[width=0.80\textwidth]{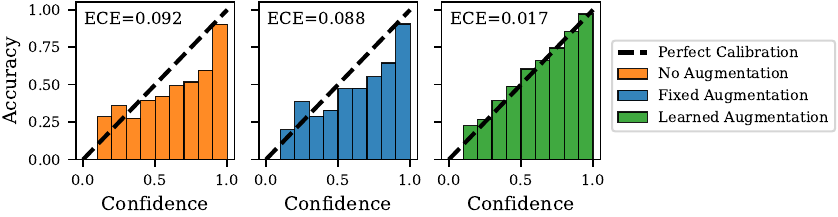}
    % %\vspace{-1ex}
    %\caption{\textbf{\ourmethod obtains the best calibration and the lowest expected calibration error (ECE)} \citep{guo2017calibration,niculescu2005predicting}. Calibration diagrams for \cifar \citep{krizhevsky2009cifar10} with ResNet-18 \citep{he2016deep}.
%    %\vspace{-1ex}
    \caption{\textbf{\ourmethod obtains the best calibration.}  Example of ResNet-18 on \cifar; see details in \cref{sec:exp_cifar10}.
    % under (\emph{left}) no augmentation, 
    % (\emph{middle}) fixed geometric augmentation, and (\emph{right}) learned geometric augmentation. 
    }
    \label{fig:cifar10_reliability}
%    %\vspace{-3.5ex}
    \vspace{-2ex}
\end{figure}

In this paper, we propose a novel approach %, which we call \ourmethod  
to optimize \da which counters these limitations. 
In particular, we take a probabilistic view of \da, whereby we treat \da parameters as model (hyper-) parameters. 
We then consider the optimization of such parameters as a Bayesian model selection problem. Due to the intractability of the Bayesian model selection objective (i.e., the marginal likelihood), we derive a tractable \ELBO, which allows us to optimize \da parameters jointly with model parameters, bypassing the need to perform expensive cross-validation or grid search. % Bayesian optimization. 
%In the case of a Bayesian treatment of model parameters, we can also derive an \ELBO to optimize an approximate distribution over model parameters jointly with \da parameters. 
%
We provide an extensive theoretical analysis, % of \ourmethod
which indicates robust predictive performance and low \ECE as demonstrated by the experiments (see, e.g., \cref{fig:cifar10_reliability}).
%
%
% Bayesian inference offers a principled approach to modeling and quantifying uncertainty in complex machine learning tasks, yet it faces challenges in large-scale settings. Data augmentation, widely adopted in frequentist practice to enrich training signals, is often applied heuristically in Bayesian scenarios, e.g., by replicating or weighting augmented samples as if they were fully independent. Such heuristics can lead to overcounting in the likelihood, inadvertently shrinking posterior uncertainty and harming calibration.
%
% We propose a novel method to \emph{learn} the data augmentation process within the Bayesian framework itself. By modeling augmentation parameters as latent variables with a distribution \(p(\mbgamma|\mbphi)\), where \(\mbphi\) itself is inferred via a variational distribution \(q(\mbphi)\), we seamlessly embed augmentation into the log-likelihood. Through Jensen's inequality and an \emph{augmented \ELBO}, we bridge the gap between na\"ive augmentation and theoretically grounded Bayesian approaches, ensuring proper marginalization and regularization of both model and augmentation parameters.
%
%\bahien{We should emphasize the robustness of \ourmethod, and the ability to optimize jointly the data augmentation together with the model.} \michael{Ref Fig. 1 in this context}
%
Our main contributions are as follows:
%\paragraph{Methodology:} We introduce an empirical Bayes perspective in which the augmentation distribution is learned directly from data, effectively performing implicit model selection over augmentation strategies.
% \paragraph{Methodology:} We introduce a novel methodological framework that frames the learning of \da parameters from an empirical Bayes perspective. This is implemented through a variational Bayes approach where both model parameters and parameters governing the augmentation distribution are learned jointly via a unified Evidence Lower BOund (ELBO). This approach allows for an implicit model selection over augmentation strategies, optimized directly from data, and it scales to complex models, moving beyond manual tuning or expensive black-box optimization. \michael{While I do like stating the contributions clearly, this paragraph could potentially be merged with the first one on this page (or trimmed down) as there is a bit of redundancy if we need to cut.}

%%\vspace{-1.8ex}

\paragraph{Methodology:} We introduce \ourmethod (OPTImizimg Marginalized Augmentations), a novel framework to learn \da parameters grounded on Bayesian principles. 
We then provide a tractable variational approximation which allows for the optimization of both model parameters and \da parameters, yielding a practical and fast alternative to manual tuning or expensive black-box optimization of \da parameters.

\paragraph{Theory:}
We provide a comprehensive theoretical analysis, establishing a cohesive framework to understand our Bayesian approach to \da, highlighting its principled nature.
Our analysis includes:
({\bf i}) The analysis of the variational approximation's quality, guiding \da distribution design (\cref{sec:variational}).
 ({\bf ii}) A derivation of PAC-Bayes generalization guarantees (\cref{sec:generalization}) and demonstration on how \ourmethod promotes model invariance and smoother decision boundaries (\cref{sec:invariance}).
 ({\bf iii}) A demonstration of improved uncertainty quantification and calibration through proper marginalization over \da parameters (\cref{sec:marginalization}).
({\bf iv}) The establishing of empirical Bayes optimality (\cref{sec:empirical-bayes}) for data-driven \da strategies, complemented by information-theoretic insights (\cref{sec:information-theory}) into how learned \da enhances inference.

\paragraph{Empirical Validation:}
We support \ourmethod and the theoretical developments with rigorous empirical validation on various tasks (\cref{sec:experiments}), including regression, image classification on standard benchmarks (e.g., \cifar and \imagenet), and an additional natural language classification task (SST-5).
Across all these settings—spanning both continuous geometric transformations and discrete text perturbations—our experiments consistently demonstrate that \ourmethod improves generalization, model calibration, and robustness to out-of-distribution data compared to models trained with fixed or no augmentation strategies.

%
%Together, these results establish \ourmethod as a principled, theoretically grounded method for integrating \da into Bayesian inference, with clear advantages over heuristic approaches.
%
Overall, our work demonstrates how Bayesian principles, specifically through a (partial or full) variational treatment of both model and augmentation parameters, can be effectively leveraged to develop a practical, scalable, and principled framework for optimizing \da, moving beyond expensive trial-and-error or validation-based procedures for optimal \da. % ad-hoc choices and unlocking more robust and reliable machine learning models.

%\maurizio{We should probably emphasize that we carry out a (partial) variational treatment of the model if this is what is done in the experiments.} \madi{There is a table where I did the experiment where only augmentation was learned (for non-bayesian nn) but I commented it since I did not compare them with fixed ones. I have a plan to do that}
% \ourmethod offers a principled remedy to heuristic \da in Bayesian deep learning, improving the reliability and interpretability of posterior predictions.

%------------------------------------------------------------------------------%

%\vspace{-1.1ex}

\section{Background and Related Work}
\label{sec:background}

%\vspace{-1.1ex}

%\maurizio{Here it looks like we will use Bayesian neural nets - we should probably emphasize this in the intro/abstract.}

We consider supervised learning tasks, where  mappings from inputs $\mbx \in \R^D$ to labels $\mby \in \R^O$ are learned from $N$ training observations $\mathcal{D} = \{(\mbx_i, \mby_i)\}_{i=1}^N$.
A common approach is to find a loss minimizing point estimates, which is equivalent to maximizing a log likelihood $\log p(\mbY \g \mbtheta, \mbX)$, where $\mbX$ and $\mbY$ denote all inputs and labels, respectively.

%\vspace{-1.5ex}

\paragraph{Marginal likelihood and \ELBO.}
In the Bayesian approach we choose a prior $p(\mbtheta)$, and infer the \textcolor{BlueViolet}{posterior distributions over parameters} and \textcolor{Bittersweet}{predictive distribution} for a new data point $\mbx^{\star}$ as:
%An alternative is to consider a Bayesian treatment of the parameters $\mbtheta$. 
%In this case, we assume a prior distribution $p(\mbtheta)$ for the parameters $\mbtheta$ and we obtain a posterior distribution over model parameters from Bayes theorem, which can then be used to obtain the predictive posterior by marginalizing parameters out:
% \begin{equation} \label{eq:bayesi:theorem}
% p(\mbtheta \g \mbX, \mbY) = \frac{p(\mbY \g \mbtheta, \mbX)p(\mbtheta)}{p(\mbY \g \mbX)}.
% \end{equation}
% The predictive posterior
% \begin{equation} \label{eq:predpost}
% p(\mby \g \mbx) = \int p( \mby | \mbx, \mbtheta) p(\mbtheta \g \mbX, \mbY) d\mbtheta
% \end{equation}
% \begin{align} \label{eq:bayesi:theorem}
%     \text{parameter posterior:}& \hspace{12mm} p(\mbtheta \g \mathcal{D}) = \frac{p(\mbY \g \mbtheta, \mbX)p(\mbtheta)}{p(\mbY \g \mbX)} \\
%     \text{predictive posterior:}& \hspace{8.5mm} p(\mby_{\mathrm{test}} \g \mbx_{\mathrm{test}}, \mathcal{D}) = \int p( \mby_{\mathrm{test}} \g \mbx_{\mathrm{test}}, \mbtheta) p(\mbtheta \g \mathcal{D}) d\mbtheta.
% \end{align}

%\vspace{-2.2ex}

% \begin{align} \label{eq:bayesi:theorem}
%     p(\mbtheta \g \mathcal{D}) = \frac{p(\mbY \g \mbtheta, \mbX)p(\mbtheta)}{p(\mbY \g \mbX)} &\qquad p(\mby^{\star}  \g \mbx^{\star}, \mathcal{D}) = \int p( \mby^{\star} \g \mbx^{\star}, \mbtheta) p(\mbtheta \g \mathcal{D}) d\mbtheta.
% \end{align}

\vspace{-1.5ex}

\begin{center}
\begin{minipage}{.4\linewidth}
  \begin{equation}
  \textcolor{BlueViolet}{p(\mbtheta \g \mathcal{D})} = \frac{p(\mbY \g \mbtheta, \mbX)p(\mbtheta)}{p(\mbY \g \mbX)}, \label{eq:bayesi:theorem}
  \end{equation}
\end{minipage}%
\begin{minipage}{.6\linewidth}
  %\vspace{1ex}
  \begin{equation}
  \textcolor{Bittersweet}{p(\mby^{\star}  \g \mbx^{\star}, \mathcal{D})} = \int p( \mby^{\star} \g \mbx^{\star}, \mbtheta) \textcolor{BlueViolet}{p(\mbtheta \g \mathcal{D})} d\mbtheta.
  \end{equation}
\end{minipage}
\end{center}

\vspace{-0.5ex}

% \begin{tabularx}{\linewidth}{XX}
% \begin{equation}
%     \hspace{-8ex} p(\mbtheta \g \mathcal{D}) = \frac{p(\mbY \g \mbtheta, \mbX)p(\mbtheta)}{p(\mbY \g \mbX)} \label{eqab}
% \end{equation}
% & 
% \begin{equation}
%     p(\mby^{\star}  \g \mbx^{\star}, \mathcal{D}) = \int p( \mby^{\star} \g \mbx^{\star}, \mbtheta) p(\mbtheta \g \mathcal{D}) d\mbtheta. \label{eqcd}
% \end{equation}
% \end{tabularx}

% marginalizes over the posterior when making predictions, thus allowing to quantification predictive uncertainty.

%Another advantage offered by a Bayesian treatment is the possibility to carry out model selection. 
The denominator of \cref{eq:bayesi:theorem} is the  \emph{marginal likelihood}, representing the data likelihood under the prior:
\begin{equation} \label{eq:evidence}
    p(\mbY \g \mbX, \mbphi) = \int p(\mbY \g \mbtheta, \mbX, \mbphi)p(\mbtheta \g \mbphi) d\mbtheta,
\end{equation}

\vspace{-0.5ex}

%\vspace{-2ex}

where we made explicit the dependence on continuous hyper-parameters $\mbphi$. We can perform model selection by 
%navigating a continuum of possible models, 
choosing the one with highest log-marginal likelihood, also known as \emph{model evidence}. The intractability of this objective motivates us to employ variational inference to obtain a tractable lower bound to be optimized with respect to a parametric surrogate posterior $q(\mbtheta)$,
%
%The marginal likelihood is typically intractable, and a popular approach is to approximate it using variational principles.
%After introducing an approximate posterior $q(\mbtheta)$, it is possible to derive the Evidence Lower-BOund: 

\vspace{-1.3ex}

\begin{align}
    \log p(\mbY \g \mbX, \mbphi) \ge \E_{q(\mbtheta)} \big[\log p(\mbY \g \mbtheta, \mbX, \mbphi)\big] - \KL\bigl[q(\mbtheta)\,\|\,p(\mbtheta \g \mbphi)\bigr] \quad =: \ELBO
\end{align}

\vspace{-1.3ex}

% \maurizio{I've used $M$ to denote a modeling choice, but we can introduce hyper-parameters $\mbphi$ and this is more consistent with the rest of the story. We can then say that these hyper-parameters can be optimized (or we can obtain a posterior), and that this is a form of model selection - selecting among a continuum of models.}

% \subsection{Bayesian Inference and Variational Methods}

% Given a prior \(p(\mbtheta)\) and data \(\mathcal{D} = \{(x_i, y_i)\}_{i=1}^N\), Bayesian inference aims to compute the posterior distribution \(p(\mbtheta|\mathcal{D})\):
% \begin{equation}
% p(\mbtheta|\mathcal{D}) = \frac{p(\mathcal{D}|\mbtheta)p(\mbtheta)}{p(\mathcal{D})}
% \end{equation}

% For complex models like neural networks, this posterior is typically intractable. Variational inference (VI) approximates it with a simpler distribution \(q(\mbtheta)\) by maximizing the Evidence Lower Bound (ELBO):
% \begin{equation}
% \ELBO(q) = \E_{q(\mbtheta)} [\log p(\mathcal{D}|\mbtheta)] - \KL\bigl(q(\mbtheta)\,\|\,p(\mbtheta)\bigr)
% \end{equation}

% Despite many algorithmic advances~\citep{blundell2015weight,gal2016dropout,louizos2017multiplicative}, applying VI to large neural networks remains challenging. Data augmentation offers a complementary approach to improve model performance, but its integration into the Bayesian framework requires careful consideration.

\paragraph{Data augmentation in neural Networks.}

In \da, we apply transformations $T_\mbgamma(\mbx)$ parameterized by $\mbgamma$ to the inputs at training time.
%, we pick a set of transformation to apply to the inputs $\mbx$ so as to augment our training set. 
In image classification common transformations include rotations, translations, flips, and color manipulations, while in natural language popular augmentations involve word substitutions and syntactic transformations \citep{shorten2019survey,feng2021survey}.
%Denote by $T_\mbgamma(\mbx)$ one of such transformations, parameterized by $\mbgamma$.  
During training, for each sample in a mini-batch, we first sample a transformation parameter $\mbgamma$ and then apply $T_\mbgamma(\mbx)$. 
This approach has proven highly effective in improving generalization in deep learning~\citep{shorten2019survey}.
%\maurizio{Comment on the success of \da thanks to regularization effects and cite the relevant literature.}

\vspace{-1.5ex}

\paragraph{Augmentations overcount evidence.}
Na\"ively replicating augmented examples \(\{(T_\mbgamma(\mbx_i), \mby_i)\}\) as if fully independent effectively multiplies the evidence \eqref{eq:evidence}, \emph{overcounting} the likelihood \citep{wilson2020bayesian}. 
For a single data point \((\mbx_i, \mby_i)\), this yields a likelihood \(\prod_{k=1}^K p(\mby_i \mid T_{\mbgamma_k}(\mbx_i),\mbtheta)\), equivalent to raising \(p(\mby_i \mid \mbx_i,\mbtheta)\) to the power \(K\).
This overcounting can artificially shrink posterior uncertainty and degrade calibration, undermining a key advantage of Bayesian methods. %\michael{Consider mentioning this in the intro to put emphasize on the calibration}
%\maurizio{Good point - and add a reference to a relevant work on temperature scaling maybe?}

% Data augmentation transforms an input \(x\) into \(T_\mbgamma(x)\) by sampling a transformation parameter \(\mbgamma\) from some distribution. Common transformations include rotations, translations, flips, and color jitter for images, or word substitutions and syntactic transformations for text.

% In standard practice, augmentation is applied by creating multiple transformed versions of each training example and treating them as additional, independent data points. This approach has proven highly effective in improving generalization in deep learning~\citep{shorten2019survey}.

% However, in Bayesian inference, na\"ively replicating augmented examples \(\{(T_\mbgamma(x_i), y_i)\}\) as if fully independent effectively multiplies the evidence, overcounting the likelihood. This overcounting can artificially shrink posterior uncertainty and degrade calibration, undermining a key advantage of Bayesian methods.

%\vspace{-1ex}

\subsection{Related Works}

\paragraph{Data augmentation optimization.}
Optimizing data augmentation (\da) parameters has been approached via computationally expensive reinforcement learning (AutoAugment \citep{cubuk2019autoaugment}), made more efficient by population-based training \citep{ho2019population}. Others formulate it as density matching, often using black-box search like Bayesian optimization \citep{snoek2012practical}, or differentiable policy search \citep{lim2019fast, cubuk2020randaugment, hataya2020faster}, or as gradient matching \citep{zheng2022deep}. Bi-level optimization has also been used, but remains expensive and often relies on strong relaxations \citep{liu2021direct, li2020differentiable, hataya2022meta, mounsaveng2021learning}. These methods typically rely on heuristics and complex search pipelines. More recently, \da has been framed as invariance-constrained learning with regularized objectives \citep{benton2020learning} or via non-parametric models solved with costly \MCMC \citep{hounie2023automatic}.

\vspace{-1.5ex}

\paragraph{Probabilistic perspectives of \da.}
Probabilistic views of \da have shown perturbed inputs can induce degenerate \citep{izmailov21posterior} or tempered likelihoods \citep{kapoor2022uncertainty}, informing studies on \da's role in the cold-posterior effect \citep{wenzel20posterior, bachmann22tempering}. \citet{nabarro22augmentation} proposed an integral likelihood similar to ours using a Jensen lower bound, and \citet{heinonen2025robust} defined an augmented likelihood via label smoothing and input mollification \citep{tran2023}. \citet{kapoor2022uncertainty} analyzed augmentations through a Dirichlet likelihood. Related latent-variable formulations also appear in work such as \citet{chen2020group} and \citet{chatzipantazis2023learning}, which consider probabilistic transformations but do not optimize augmentation parameters within a joint Bayesian model. However, these approaches generally use fixed, unoptimized augmentation parameters. In contrast, \citet{wang23BA} modeled \da with stochastic output layers and auxiliary variables for MAP optimization via expectation maximization, while \citet{wu24} applied MixUp \citep{zhang2017mixup} for martingale posteriors \citep{fong2023martingale}. Broader connections link \da to kernel methods for task-specific invariances \citep{dao2019kernel}, though not directly to Bayesian inference. More directly, \citet{mark2018invariance} learned invariances via marginal likelihood for Gaussian processes \citep{williams2006gaussian}, an idea \citet{immer2022invariance} extended to BNNs \citep{neal1996bayesian,tran2022bnnprior} using the Laplace approximation \citep{mackay1992practical,daxberger2021laplace,immer2021scalable}, but without theoretical generalization guarantees.

\paragraph{PAC-Bayes generalization bounds.}
PAC-Bayes bounds \citep{mcallester1999pac, catoni2007pac, alquier2024user} offer theoretical guarantees for Bayesian methods, including in deep learning \citep{dziugaite2017computing, lotfi2022pacbayes, wilson2025deep}. However, prior work rarely treats augmentation parameters as latent variables within this framework. We unify these directions by making the augmentation distribution a key component of the model's likelihood, deriving novel theoretical results that characterize its benefits.
% {\textcolor{red}{
% \cite{chatzipantazis2023learning} approaches DA through transformed risk minimization and uses PAC-Bayes theory to derive a regularizer that prevents collapse to trivial augmentations when optimizing empirical transformed risk. In contrast, our work takes a fundamentally different Bayesian model selection approach, treating augmentation parameters as model (hyper-)parameters within a fully probabilistic framework.
% }}

% \bahien{We could discuss about "Learning Augmentation Distributions using Transformed Risk Minimization" as suggested by an NeurIPS reviewer.}

%------------------------------------------------------------------------------%
\section{Augmentation Optimization through Bayesian Model Selection}
\label{sec:method}

%\vspace{-1ex}

% \maurizio{I suggest using the notation with bold-faced vectors and matrices - similar to what I wrote in sec 2}

%In this section we take a probabilistic view of \da, which allows us to cast the optimization of \da parameters as Bayesian model selection. 
% In this section we treat the optimization of \da parameters as Bayesian model selection. 
%This section presents \ourmethodology, contrasts it with na\"ive approaches, and provides an efficient algorithm for implementation.

% We propose a principled approach to integrate \da into Bayesian inference by treating augmentation parameters as latent variables with learnable distributions. This section presents \ourmethodology, contrasts it with na\"ive approaches, and provides an efficient algorithm for implementation.

\paragraph{Augmentation as Marginalization.}

In this section we treat the optimization of \da parameters as Bayesian model selection.
To do so, we start by defining a transformation-augmented likelihood:
%We incorporate transformations directly into the likelihood. For each example \((\mbx, \mby)\), we define:
\begin{equation}
p(\mby \g \mbx, \mbtheta, \mbphi) = \E_{p(\mbgamma \g \mbphi)} \Big[ p\big(\mby \g T_\mbgamma(\mbx), \mbtheta\big) \Big],
\end{equation}

%\vspace{-2ex}

where $T_\mbgamma(\mbx)$ is the transformed input under augmentation distribution $\mbgamma \sim p(\mbgamma \g \mbphi)$ parameterized by $\mbphi$. This formulation treats augmentation as marginalization over transformations rather than data replication.
This method averages over transformations to contribute each original example exactly once, as opposed to the overcounting effect in the case of na\"ive augmentation. As we will see, this yields a more calibrated posterior with appropriate uncertainty quantification.

%\vspace{-0.5ex}

The data likelihood given model parameters \(\mbtheta\) and augmentation parameters \(\mbphi\) is

\vspace{-1.7ex}

\begin{equation}
    p(\mathcal{D} \g \mbtheta, \mbphi) = \prod_{i=1}^N \E_{p(\mbgamma \g \mbphi)} \Big[p(\mby_i \g T_\mbgamma(\mbx_i), \mbtheta)\Big].
\end{equation}

\vspace{-1.5ex}

Taking a fully Bayesian treatment, we assign a prior \(p(\mbphi)\) on the augmentation parameters \(\mbphi\), making \(\mbphi\) a latent variable alongside \(\mbtheta\). The joint distribution over all variables is $p(\mathcal{D}, \mbtheta, \mbphi, \mbgamma) = p(\mbtheta) p(\mbphi) p(\mbgamma \g \mbphi) p(\mathcal{D} | \mbtheta, \mbphi)$.
% \begin{equation}
%     p(\mathcal{D}, \mbtheta, \mbphi, \mbgamma) = p(\mbtheta) p(\mbphi) p(\mbgamma \g \mbphi) \prod_{i=1}^N p(\mby_i \g T_\mbgamma(\mbx_i), \mbtheta).
% \end{equation}
%
Our goal is to approximate the posterior \(p(\mbtheta, \mbphi \g \mathcal{D})\), which is typically intractable.
To address this challenge, we employ variational inference \citep{jordan1999introduction}.

%\vspace{-1.1ex}

\paragraph{Augmented Evidence Lower Bound.}
% \label{subsec:augmented-elbo}

For variational inference, we introduce a variational distribution \(q(\mbtheta, \mbphi) = q(\mbtheta) q(\mbphi)\) to approximate the posterior \(p(\mbtheta, \mbphi \g \mathcal{D})\). The standard \ELBO is a lower bound on the log marginal likelihood \( \mathcal{L} := \log p(\mathcal{D}) = \log \iiint p(\mathcal{D}, \mbtheta, \mbphi, \mbgamma) \, d\mbtheta \, d\mbphi \, d\mbgamma \).
% \begin{align}
% \log p(\mathcal{D}) = \log \int \int \int p(\mathcal{D},\mbtheta,\mbphi, \mbgamma) \, d\mbtheta \, d\mbphi \, d\mbgamma.    
% \end{align}
Using Jensen’s inequality with \(q(\mbtheta,\mbphi)\) and with standard manipulations, we obtain the \ELBO, which consists of a \textcolor{BlueViolet} {data-fitting term} and two regularization terms  \textcolor{Bittersweet}{\(\text{KL}(q(\mbtheta) \| p(\mbtheta))\)}  and \textcolor{purple}{\(\text{KL}(q(\mbphi) \| p(\mbphi))\)}:

\vspace{-1.5ex}

\begin{equation}
    % \log p(\mathcal{D}) 
    \mathcal{L} \geq  {\textcolor{BlueViolet}{
    %\mathbb{E}_{q(\mbtheta)} \mathbb{E}_{q(\mbphi)} \mathbb{E}_{p(\mbgamma \g\mbphi)
    \mathbb{E}_{q(\mbtheta) q(\mbphi) p(\mbgamma \g\mbphi)
    } 
    \left[ \sum_{i=1}^N \log p(\mby_i \g T_\mbgamma(\mbx_i),\mbtheta) \right]}} - {\textcolor{Bittersweet}{\text{KL}(q(\mbtheta) \| p(\mbtheta))}} - {\textcolor{purple}{\text{KL}(q(\mbphi) \| p(\mbphi))}}. \label{eq:aug_elbo_def}
\end{equation}

\vspace{-1.5ex}

\paragraph{Optimization of the \ELBO.}

%We optimize the augmented ELBO by alternating gradient steps on the parameters of \(q(\mbtheta)\) and \(q(\mbphi)\). This approach ensures stable optimization, as \(\mbtheta\) and \(\mbphi\) may operate on different scales or require distinct learning rates, a common strategy in variational inference (VI) \citep{jordan1999introduction} with multiple parameter sets~\citep{kingma2014adam}. Alternating updates also enhance convergence by preventing one parameter set from dominating the optimization process.

\label{subsec:optimization_strategy} % Or a more descriptive label
The augmented \ELBO presented in \cref{eq:aug_elbo_def} is optimized by jointly updating the parameters of the variational distributions \(q(\mbtheta)\) and \(q(\mbphi)\) using stochastic gradient-based methods. This involves sampling from these distributions (often via reparameterization) and from the \da distribution \(p(\mbgamma \g \mbphi)\) to compute Monte Carlo estimates of the expectation term, and then backpropagating through the objective. A detailed algorithm, specific choices for parameterizing \(p(\mbgamma \g \mbphi)\) and \(q(\mbphi)\) for continuous and discrete transformations, and other practical implementation aspects are discussed in \cref{sec:algorithm}. %The code is available in \url{https://github.com/imadik31/OPTIMA}.

\vspace{-1.3ex}

%------------------------------------------------------------------------------%
\section{Theoretical Analysis}
\label{sec:theory}

%\vspace{-1.8ex}

We present a comprehensive %theoretical framework for characterizing the properties of the proposed 
analysis of the proposed \da approach based on Bayesian model selection, analyzing its properties from multiple perspectives: variational approximation quality, generalization guarantees, invariance properties, and connections to empirical Bayes. 
Our analysis includes a direct comparison with na\"ive \da, which amounts in treating augmented samples as training samples.
This analysis provides a rigorous foundation for \ourmethod while yielding practical insights for implementation.

%\vspace{-1.8ex}

\subsection{Variational Approximation with Augmentation}
\label{sec:variational}

We begin by analyzing the quality of our variational approximation when incorporating \da.
% We first quantify how closely our variational formulation approximates the true marginalization, by bounding the Jensen gap introduced by moving the expectation inside the log-likelihood.

\begin{proposition}[Jensen Gap Bound]
\label{prop:jensen-gap}
The augmentation distribution variance and model sensitivity control the Jensen gap introduced by our lower bound approximation. If \( f(\mbgamma) = \log p(\mby \mid T_\mbgamma(\mbx),\mbtheta) \) is \(L\)-Lipschitz in \(\mbgamma\), and \(\mbgamma \sim p(\mbgamma|\mbphi)\) is sub-Gaussian with variance proxy \(\sigma^2\), then:

\vspace{-2.2ex}

\begin{align}
\log \E_{\mbgamma}\bigl[p(\mby \g T_\mbgamma(\mbx),\mbtheta)\bigr] - \E_{\mbgamma}\bigl[\log p(\mby \g T_\mbgamma(\mbx),\mbtheta)\bigr] \leq \frac{L^2 \sigma^2}{2}.    
\end{align}

\vspace{-1.8ex}

Also, this bound is tight when \(f(\mbgamma)\) is approximately linear in the high-probability region of \(p(\mbgamma|\mbphi)\).
\end{proposition}

%\vspace{-1.5ex}

The proof is presented in \cref{{sec:proof_prop_jensen-gap}}.
This result has important implications for optimizing the augmentation distribution \(p(\mbgamma|\mbphi)\):

\begin{mybox}
\begin{corollary}[Optimal Augmentation Variance]
\label{cor:optimal-variance}
The optimal variance \(\sigma^2_\mbphi\) for the augmentation distribution balances two competing factors:
%\vspace{-0.6ex}
\begin{enumerate}
    \item Increasing \(\sigma^2_\mbphi\) improves exploration of the augmentation space.
%\vspace{-0.6ex}
    \item Decreasing \(\sigma^2_\mbphi\) tightens the variational bound.
\end{enumerate}
%\vspace{-0.8ex}
For models with high sensitivity to augmentations (large \(L\)), smaller variance is preferred to maintain bound tightness.
\end{corollary}
    
\end{mybox}

%\vspace{-0.9ex}

This corollary provides practical guidance for setting augmentation distribution parameters, suggesting that highly sensitive models benefit from more conservative augmentation strategies.

%\vspace{-1.5ex}

\subsection{Generalization Guarantees}
\label{sec:generalization}

%\vspace{-1.5ex}

To analyze the generalization of our Bayesian-optimized \da, we leverage the PAC-Bayes framework \citep{mcallester1999pac,catoni2007pac}; see \cref{app:pac-bayes-primer} for a primer. 
PAC-Bayes theory provides high-probability upper bounds on the true risk (generalization error) of a learning algorithm that outputs a distribution over hypotheses (a ``posterior''). These bounds typically depend on the empirical risk observed on the training data and a complexity term, often expressed as the KL divergence between this posterior and a data-independent prior distribution. By extending this framework to our setting, we can formally quantify how well the model with learned \da parameters will perform on unseen data. We first present a PAC-Bayes bound for \ourmethod, then provide a theorem that explicitly compares \ourmethod to na\"ive \da, demonstrating superior generalization.

\begin{definition}[True and Empirical Risks]
Given the transformation function $T_\mbgamma(\mbx)$, we define:

%\vspace{-2ex}

\begin{itemize}
    % \item \emph{True risk:} \(R(\mbtheta,\mbphi) = \mathbb{E}_{(\mbx,\mby) \sim \mathcal{D}} \left[ -\log \mathbb{E}_{p(\mbgamma \mid\mbphi)} p(\mby \mid T_\mbgamma(\mbx),\mbtheta) \right]\).

    \item \emph{True risk:} \(R(\mbtheta,\mbphi) = \mathbb{E}_{(\mbx,\mby) \sim P} \left[ -\log \mathbb{E}_{p(\mbgamma \mid\mbphi)} p(\mby \mid T_\mbgamma(\mbx),\mbtheta) \right]\).

    %\vspace{-1.1ex}
    
    \item \emph{Our empirical risk:} \(\hat{R}(\mbtheta,\mbphi) = -\frac{1}{N} \sum_{i=1}^N \log \mathbb{E}_{p(\mbgamma \mid\mbphi)} p(\mby_i \mid T_\mbgamma(\mbx_i),\mbtheta)\).

    %\vspace{-1.1ex}

    \item \emph{Empirical risk for na\"ive augmentation ($K$ samples per datapoint)}: 

    \vspace{-2.5ex}
    
    \[\hat{R}_{\text{na\"ive}}(\mbtheta) = -\frac{1}{N} \frac{1}{K} \sum_{i=1}^N \sum_{k=1}^K \log p(y_i \mid T_{\mbgamma_k}(x_i),\mbtheta), \quad \mbgamma_k \sim p(\mbgamma \mid\mbphi)
    \]
    
\end{itemize}
\end{definition}

%\vspace{-1.5ex}

% We next study how our augmented likelihood interacts with standard PAC-Bayes theory, yielding a generalization bound for OPTIMA.

\begin{theorem}[PAC-Bayes with Augmented Likelihood]
\label{thm:pac-bayes}
% For an i.i.d. dataset \(\mathcal{D} = \{(\mbx_i, \mby_i)\}_{i=1}^N\) drawn from an unknown distribution, any prior
For an i.i.d. dataset $\mathcal{D} = \{(\mbx_i, \mby_i)\}_{i=1}^N$ drawn from an unknown distribution $P$, any prior \(p(\mbtheta,\mbphi)\), and any posterior \(q(\mbtheta,\mbphi) = q(\mbtheta) q(\mbphi)\) over the hypothesis space \(\mbtheta \times\mbphi\), with probability at least \(1 - \delta\) over the draw of \(\mathcal{D}\):
\begin{align}
    %\vspace{-1.5ex}
    \mathbb{E}_{q(\mbtheta,\mbphi)} \big[R(\mbtheta,\mbphi)\big] \leq \mathbb{E}_{q(\mbtheta,\mbphi)} \big[\hat{R}(\mbtheta,\mbphi)\big] + \sqrt{\frac{\KL(q(\mbtheta,\mbphi) \| p(\mbtheta,\mbphi)) + \log\frac{2\sqrt{N}}{\delta}}{2N}},
\end{align}

%\vspace{-1.5ex}

where \(\text{KL}(q(\mbtheta,\mbphi) \| p(\mbtheta,\mbphi)) = \text{KL}(q(\mbtheta) \| p(\mbtheta)) + \text{KL}(q(\mbphi) \| p(\mbphi))\) if \(p(\mbtheta,\mbphi) = p(\mbtheta) p(\mbphi)\).
\end{theorem}

To explicitly demonstrate that \ourmethod generalizes better than na\"ive \da, we now compare the PAC-Bayes bounds of both methods, showing that \ourmethod yields a tighter bound due to proper marginalization over transformations.
% \textcolor{red}{
We encourage reader refer to \cref{app:pac-bayes-primer} for further discussion.
% }

% \bahien{We could point out more clearly which are new results compared to standard PAC-Bayes as suggested by Reviewer H1Vj.}
% \noindent\textbf{Remark.}
% As is standard when replacing expectations by Monte Carlo averages, the empirical naïve-augmentation risk $\hat{R}_{\mathrm{naive}}$ provides a consistent approximation of the true marginalization when $K$ is sufficiently large. We make this implicit assumption in Theorem~\ref{thm:generalization-advantage}
\noindent\textbf{Remark.}
As usual with Monte Carlo estimates, the naïve risk $\hat{R}_{\mathrm{naive}}$ is a consistent approximation of the true marginalization when $K$ is large enough, which is the setting assumed in Theorem~\ref{thm:generalization-advantage}.

\begin{theorem}[Generalization Advantage of Bayesian-Optimized Augmentation]
\label{thm:generalization-advantage}
% Let \(\mathcal{D} = \{(x_i, y_i)\}_{i=1}^N\) be an i.i.d. dataset drawn from a distribution \(\mathcal{D}\).
% Consider a model parameterized by \(\mbtheta \in\mbTheta\), and let \(\mbphi \in\mbphi\) parameterize an augmentation distribution \(p(\mbgamma \mid\mbPhi)\), where \(\mbgamma\) defines transformations \(T_\mbgamma(\mbx)\).
Consider a model parameterized by $\mbtheta \in \Theta$, and let $\mbphi \in \Phi$
parameterize an augmentation distribution $p(\mbgamma \mid \mbphi)$, where $\mbgamma$
defines transformations $T_\mbgamma(\mbx)$.

We consider the following assumptions:
\begin{enumerate}[leftmargin=7mm]

    %\vspace{-1.3ex}

    \item The transformation distribution \(p(\mbgamma \mid\mbphi)\) is such that \(\mathbb{E}_{p(\mbgamma \mid\mbphi)} p(\mby \mid T_\mbgamma(\mbx),\mbtheta)\) can be computed or approximated accurately. 

    %\vspace{-1.2ex}
    
    \item The variational posteriors \(q(\mbtheta,\mbphi)\) and \(q(\mbtheta)\) are optimized to minimize their respective bounds.

   %\vspace{-1.1ex}
    
    \item The KL divergences \(\text{KL}(q(\mbtheta,\mbphi) \| p(\mbtheta,\mbphi))\) and \(\text{KL}(q(\mbtheta) \| p(\mbtheta))\) are comparable, i.e., the complexity penalties are similar.
\end{enumerate}

% \textbf{Definitions:}
% \begin{itemize}
%     \item \textbf{True Risk (\ourmethod):}
%     \[
%     R(\mbtheta,\mbphi) = \mathbb{E}_{(x,y) \sim \mathcal{D}} \left[ -\log \mathbb{E}_{p(\mbgamma \mid\mbphi)} p(y \mid T_\mbgamma(x),\mbtheta) \right]
%     \]
%     \item \textbf{Empirical Risk (\ourmethod):}
%     \[
%     \hat{R}(\mbtheta,\mbphi) = -\frac{1}{N} \sum_{i=1}^N \log \mathbb{E}_{p(\mbgamma \mid\mbphi)} p(y_i \mid T_\mbgamma(x_i),\mbtheta)
%     \]
%     \item \textbf{Empirical Risk (na\"ive Augmentation):} For \(K\) augmentations per data point,
%     \[
%     \hat{R}_{\text{na\"ive}}(\mbtheta) = -\frac{1}{N} \sum_{i=1}^N \frac{1}{K} \sum_{k=1}^K \log p(y_i \mid T_{\mbgamma_k}(x_i),\mbtheta), \quad \mbgamma_k \sim p(\mbgamma \mid\mbphi)
%     \]
% \end{itemize}

% \textbf{Assumptions:}

% \textbf{Theorem Statement:}

%\vspace{-1.6ex}

Under these assumptions, the PAC-Bayes bound for \ourmethod is tighter than that for na\"ive \da: %, i.e.,
\begin{align}
\mathbb{E}_{q(\mbtheta,\mbphi)} [R(\mbtheta,\mbphi)] \leq \mathbb{E}_{q(\mbtheta)} [R(\mbtheta)] - \Delta,
\end{align}
where \(\Delta = \mathbb{E}_{q(\mbtheta,\mbphi)} \left[ \frac{1}{N} \sum_{i=1}^N \Delta_{\mbphi}(\mbx_i, \mby_i) \right] \geq 0\), and \(\Delta_{\mbphi}(\mbx_i, \mby_i) = \log \mathbb{E}_{p(\mbgamma \mid\mbphi)} p(\mby_i \mid T_\mbgamma(\mbx_i),\mbtheta) - \mathbb{E}_{p(\mbgamma \mid\mbphi)} \log p(\mby_i \mid T_\mbgamma(\mbx_i),\mbtheta)\).
Furthermore, \(\Delta > 0\) when \(p(\mby_i \mid T_\mbgamma(\mbx_i),\mbtheta)\) varies across \(\mbgamma\), indicating a strictly better generalization bound for \ourmethod.

%\vspace{-1.8ex}

\end{theorem}

The proofs for \cref{thm:pac-bayes} and \cref{thm:generalization-advantage} are detailed in \cref{sec:proof_thm_pac_bayes} and \cref{sec:proof_thm_generalization_advantage}. This theorem provides several key insights:

%\vspace{-0.3ex}

\begin{mybox}
    
\begin{corollary}[Marginalization Advantage]
\label{cor:marginalization-advantage}
For a fixed \(\mbphi\), the term \(\Delta_{\mbphi}(\mbx_i, \mby_i) = \log \mathbb{E}_{p(\mbgamma \mid\mbphi)} p(\mby_i \mid T_\mbgamma(\mbx_i),\mbtheta) - \mathbb{E}_{p(\mbgamma \mid\mbphi)} \log p(\mby_i \mid T_\mbgamma(\mbx_i),\mbtheta)\) quantifies the advantage of proper marginalization over na\"ive \da. By Jensen's inequality, \(\Delta_{\mbphi}(\mbx_i, \mby_i) \geq 0\), with equality only when \(p(\mby_i \mid T_\mbgamma(\mbx_i),\mbtheta)\) is constant across all \(\mbgamma\) in the support of \(p(\mbgamma|\mbphi)\).
\end{corollary}

\end{mybox}
% This immediately yields a quantifiable advantage of marginalization over naïve augmentation.
%\vspace{-1ex}

\begin{mybox}
\begin{corollary}[Augmentation-Aware Prior]
\label{cor:augmentation-prior}
% The generalization bound is minimized when the prior \(p(\mbtheta,\mbphi)\) encodes invariance to the transformations in the augmentation family. Specifically, if \(p(\mbtheta)\) assigns high probability to parameters that satisfy \(p(\mby \mid T_\mbgamma(\mbx),\mbtheta) \approx p(\mby \mid \mbx,\mbtheta)\) for all \(\mbgamma\), and \(p(\mbphi)\) favors augmentation distributions that align with the data’s structure, then the \(\text{KL}(q(\mbtheta,\mbphi) \| p(\mbtheta,\mbphi))\) term will be smaller.
The PAC-Bayes bound is smallest when the prior \(p(\mbtheta,\mbphi)\) reflects the invariances induced by the augmentation family. Priors that favor parameters satisfying \(p(\mby \mid T_\mbgamma(\mbx),\mbtheta)\approx p(\mby \mid \mbx,\mbtheta)\) lead to smaller KL terms and tighter bounds.
\end{corollary}
\end{mybox}

These results demonstrate that \ourmethod provides better generalization guarantees than na\"ive \da and suggests principles for designing priors that complement the \da strategy.

%\vspace{-1.5ex}

\subsection{Invariance Analysis}
\label{sec:invariance}

%\vspace{-1.5ex}

We now analyze how \ourmethod promotes invariance to transformations, extending beyond first-order (Jacobian-based) analysis to include higher-order effects. This analysis reveals how the model’s sensitivity to input transformations is regularized, encouraging robustness and generalization.

%\vspace{-0.1ex}

\begin{theorem}[Higher-Order Invariance]
\label{thm:invariance}
Let \( f_\mbtheta \) be a twice-differentiable function parameterized by \(\mbtheta \), with its Hessian bounded such that \( \|\nabla^2 f_\mbtheta\| \leq H \). For input transformations \( T_\mbgamma(\mbx) = x + \delta(\mbgamma) \), where \( \delta(\mbgamma) \) is a perturbation with zero mean, \( \mathbb{E}_{p(\mbgamma|\mbphi)}[\delta] = 0 \), and covariance \( \mathbb{E}_{p(\mbgamma|\mbphi)}[\delta \delta^{\top}] = \Sigma_\mbphi \), the expected squared difference in the model’s output under these transformations is:
\begin{align}
\mathbb{E}_{p(\mbgamma|\mbphi)} \left[ \| f_\mbtheta(T_\mbgamma(\mbx)) -  f_\mbtheta(\mbx) \|^2 \right] &= \textcolor{Bittersweet}{\mathrm{Tr} \left( J_f(\mbx)^{\top} J_f(\mbx) \Sigma_\mbphi \right)} \nonumber \\
&+ \textcolor{purple}{\frac{1}{4} \mathbb{E}_{p(\mbgamma|\mbphi)} \left[ \delta^{\top} \nabla^2 f_\mbtheta(\mbx)^{\top} \nabla^2 f_\mbtheta(\mbx) \delta \right]} +  \textcolor{BlueViolet}{\mathcal{O}(\|\delta\|^3)},
% &\quad + \frac{1}{4} \mathbb{E}_{p(\mbgamma|\mbphi)} \left[ \delta^{\top} \nabla^2 f_\mbtheta(\mbx)^{\top} \nabla^2 f_\mbtheta(\mbx) \delta \right] \\
% &\quad + \mathcal{O}(\|\delta\|^3),
\end{align}

%\vspace{-1.5ex}

where \( J_f(\mbx) \) is the Jacobian of \( f_\mbtheta \) at input \( x \), \( \nabla^2 f_\mbtheta(\mbx) \) is the Hessian of \( f_\mbtheta \) at \( x \), and \( \textcolor{BlueViolet}{\mathcal{O}(\|\delta\|^3)} \) represents higher-order terms that become negligible for small perturbations.
% \begin{itemize}
%     \item 
%     \item 
%     \item \( \mathrm{Tr}(\cdot) \) denotes the trace of a matrix,
%     \item 
% \end{itemize}
\end{theorem}

%\vspace{-1.0ex}

\begin{mybox}
\begin{corollary}[Input-Space Regularization]
\label{cor:flat-minima}
The second-order term in \cref{thm:invariance} acts as a regularizer, penalizing high curvature in the model’s output with respect to the input. This encourages a smoother response surface, promoting robustness to transformations and potentially enhancing generalization by reducing sensitivity to irrelevant input variations.
\end{corollary}
\end{mybox}

%\vspace{-1ex}

% This result reveals how \da regularizes the model through both first-order (sensitivity) and second-order (curvature) terms.
% The first-order term \( \textcolor{Bittersweet}{\mathrm{Tr} \left( J_f(\mbx)^{\top} J_f(\mbx) \Sigma_\mbphi \right)} \) captures the model’s local sensitivity to input perturbations.
% The second-order term \( \textcolor{purple}{\frac{1}{4} \mathrm{Tr} \left( \nabla^2 f_\mbtheta(\mbx)^{\top} \nabla^2 f_\mbtheta(\mbx) \Sigma_\mbphi \right) }\) penalizes curvature, encouraging a flatter response surface in the directions defined by \( \Sigma_\mbphi \).
% By minimizing the expected squared difference as part of the augmented \ELBO, \ourmethod promotes models that are robust to transformations, with smoother outputs that generalize better to unseen data.

%\vspace{-0.6ex}

\begin{mybox}
\begin{corollary}[Optimal Transformation Covariance]
\label{cor:optimal-covariance}
The optimal covariance structure \(\Sigma_\mbphi\) for the augmentation distribution depends on the geometry of the model's response surface. Specifically, \(\Sigma_\mbphi\) should allocate more variance in directions where the model is approximately invariant (small eigenvalues of \( J_f(\mbx)^{\top} J_f(\mbx) \)) and less variance in directions of high sensitivity.
\end{corollary}
\end{mybox}

\begin{remark}
In our framework, \(\mbphi\) is inferred via \(q(\mbphi)\) by maximizing the augmented \ELBO, allowing the \da distribution to adapt to the data and further enhance model robustness.
\end{remark}

%\vspace{-0.6ex}

It provides practical guidance for designing augmentation distributions that align with the model’s natural invariances, enhancing robustness and generalization. The proof can be found in \cref{sec:proof_thm_invariance}.

%\vspace{-1.5ex}

\subsection{Marginalization vs. Heuristic Augmentation}
\label{sec:marginalization}

%\vspace{-0.5ex}

We now quantify the difference between our marginalization approach and na\"ive \da, focusing on the impact on posterior uncertainty.
The next theorem assumes local Gaussianity of the posterior with full-rank covariance, which might not be the case in practice for over-parameterized models; however, we believe that the theoretical development gives some useful insights into the behavior of \ourmethod compared to na\"ive \da, and we will attempt a more general proof in the future. 

\begin{theorem}[Posterior Shrinkage under Na\"ive Augmentation]
\label{thm:posterior-shrinkage}
Let \(p_{\text{true}}(\mbtheta \g \cD)\) be the posterior under our marginalization approach and \(p_{\text{na\"ive}}(\mbtheta \g \cD)\) be the posterior under na\"ive \da with \(K\) augmentations per data point. Under regularity conditions and assuming a locally Gaussian approximation around the MAP estimate \(\hat{\mbtheta}\):
% \begin{align}
% \Sigma_{\text{na\"ive}} \approx \frac{1}{K} \Sigma_{\text{true}},
% \end{align}
$\Sigma_{\text{na\"ive}} \approx \frac{1}{K} \Sigma_{\text{true}}$, 
where \(\Sigma_{\text{na\"ive}}\) and \(\Sigma_{\text{true}}\) are full-rank covariance matrices of \(p_{\text{na\"ive}}(\mbtheta \g \cD)\) and \(p_{\text{true}}(\mbtheta \g \cD)\), respectively.

%\vspace{-1.5ex}

\end{theorem}

The proof is in \cref{sec:proof_thm_posterior_shrinkage}. This result has significant implications for uncertainty quantification:

%\vspace{-1.1ex}

% \begin{mybox}

% \begin{corollary}[Calibration Error under na\"ive Augmentation]
% \label{cor:calibration-error}
% The expected calibration error (ECE) under na\"ive augmentation scales approximately with \(\sqrt{K-1}\), where \(K\) is the number of augmentations per data point.
% \end{corollary}

% \end{mybox}

%\vspace{-1.0ex}

\begin{mybox}

\begin{corollary}[Uncertainty Propagation]
\label{cor:uncertainty-propagation}
Predictive uncertainty is underestimated by a factor of approximately \(\sqrt{K}\) under na\"ive augmentation, leading to overconfident predictions, particularly for out-of-distribution inputs.
\end{corollary}

\end{mybox}

%\vspace{-0.1ex}

These results provide a quantitative characterization of the benefits of proper marginalization over na\"ive augmentation, particularly for uncertainty quantification and calibration.

%\vspace{-1.0ex}

\subsection{Empirical Bayes Perspective}
\label{sec:empirical-bayes}

%\vspace{-0.5ex}

Finally, we analyze \ourmethod from an empirical Bayes perspective \citep{robbins1992empirical, efron2012large}, showing how it naturally leads to optimal augmentation strategies.

\begin{theorem}[Empirical Bayes Optimality via Augmented \ELBO]
\label{thm:emp-bayes}
The augmented $\mathrm{ELBO}_{\text{aug}}(q_\mbtheta, q_\mbphi)$ (see \cref{eq:aug_elbo_def}) holds when \(q(\mbtheta) = p(\mbtheta \g \mathcal{D})\) and \(q(\mbphi) = p(\mbphi \g \mathcal{D})\). Consequently, maximizing \(\mathrm{ELBO}_{\text{aug}}(q_\mbtheta, q_\mbphi)\) with respect to both \(q(\mbtheta)\) and \(q(\mbphi)\) approximates the posterior distributions \(p(\mbtheta \g \mathcal{D})\) and \(p(\mbphi \g \mathcal{D})\), with the mode or mean of \(q(\mbphi)\) serving as a point estimate analogous to an Empirical Bayes solution, regularized by the prior \(p(\mbphi)\).
\end{theorem}

\begin{mybox}
    
\begin{corollary}[Data-Driven Augmentation]
\label{cor:data-drive-aug}
The optimization of \(q(\mbphi)\) via the augmented \ELBO results in an augmentation distribution \(p(\mbgamma \g\mbphi)\) that is specifically tailored to the observed data \(\mathcal{D}\), with \(\mbphi \sim q(\mbphi)\). This process effectively selects \da parameters enhancing the ability of the model to explain the data, implicitly performs model selection over the space of augmentation strategies.
\end{corollary}
\end{mybox}

%\vspace{-1.0ex}

\begin{mybox}

\begin{corollary}[Convergence of Joint Optimization]
Under mild regularity conditions (e.g., continuity and boundedness of the likelihood and prior), the alternating optimization of the variational distributions \(q(\mbtheta)\) and \(q(\mbphi)\) converges to a local optimum of the marginal likelihood \(p(\mathcal{D})\). This ensures that the learned augmentation distribution \(p(\mbgamma \g\mbphi)\) is both data-consistent and aligned with the model’s posterior distribution.
\end{corollary}
\end{mybox}

These results establish \ourmethod as a principled, data-driven method for learning optimal augmentation strategies within a Bayesian framework. The proof is detailed in \cref{sec:proof_thm_emp_bayes}.

\subsection{Information-Theoretic Perspective}
\label{sec:information-theory}

We now provide an information-theoretic analysis of \ourmethod, offering additional insights into the role of \da in Bayesian inference.

\begin{theorem}[Information Gain from Augmentation]
\label{thm:information-gain}
The expected information gain from \da, measured as the reduction in posterior entropy, is:
\begin{align}
\Delta H = H[p(\mbtheta \g \mathcal{D}_{\mathrm{no aug}})] - H[p(\mbtheta \g \mathcal{D})] \approx \frac{1}{2}\log\det(I + H_{\mathrm{no aug}}^{-1} H_{\mathrm{aug}}),
\end{align}
where \(H_{\mathrm{no aug}}\) and \(H_{\mathrm{aug}}\) are the Hessians of the negative log-likelihood without and with \da, respectively, and \(p(\mbtheta \g \mathcal{D})\) uses the marginalized likelihood.
\end{theorem}

The proof is in \cref{sec:proof_thm_info_gain}. This information-theoretic perspective provides additional insights:

\begin{mybox}   
\begin{corollary}[Optimal Information Gain]
\label{cor:optimal-info}
The \da distribution that maximizes information gain while maintaining a fixed KL divergence from a reference distribution aligns with the eigenvectors of the Fisher information matrix, with variance inversely proportional to the eigenvalues.
\end{corollary}
\end{mybox}

%\vspace{-1.0ex}

\begin{mybox}
\begin{corollary}[Connection to Information Bottleneck]
\label{cor:info-bottleneck}
\ourmethod can be viewed as implementing an information bottleneck, where the \da distribution \(p(\mbgamma|\mbphi)\) is optimized to maximize the mutual information between the augmented inputs and the targets, while minimizing the mutual information between the original and augmented inputs.
\end{corollary}    
\end{mybox}

These information-theoretic results provide a complementary perspective on \ourmethod, connecting it to principles of optimal experimental design and information bottleneck theory.

%------------------------------------------------------------------------------%

%------------------------------------------------------------------------------%

\vspace{-1.5ex}

\section{Experiments}
\label{sec:experiments}

\vspace{-1ex}

\subsection{Synthetic regression example}

% We evaluate our Bayesian-optimized \da approach on a toy regression problem and image classification tasks involving the \cifar % \citep{krizhevsky2009cifar10} 
% and \imagenet % \citep{deng2009imagenet} 
% datasets. 
% We begin with a toy regression problem to demonstrate the benefits of \ourmethod over na\"ive \da, (\textbf{Na\"ive Aug}), fixed \da ({\bf Fixed Aug}) or no augmentation ({\bf No Aug}).

%\bahien{It would be great if we could shorten the experimental settings, just highlight the important points. The more details could be in Appendix.}

%\vspace{-1.7ex}

% \paragraph{Illustration on Synthetic Regression.}
\label{sec:exp_synthetic_complex}

\begin{figure}[H]
    \centering
    % Adjust the width or path to match your setup
    \includegraphics[width=\textwidth]{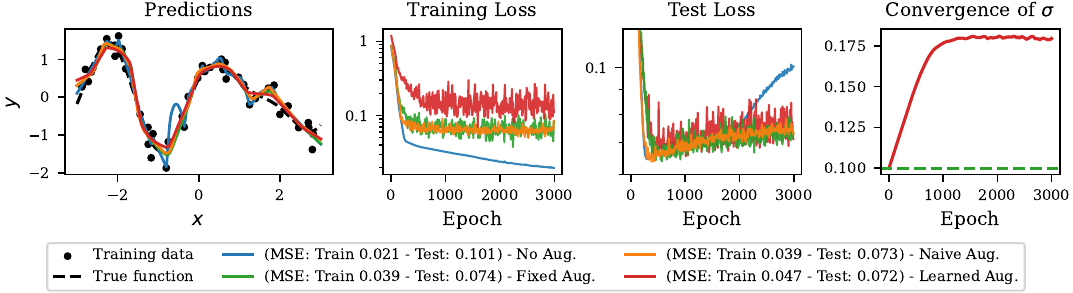}
    \vspace{-1.2ex}
    \caption{
    %\maurizio{Make plots smaller (fit four in one row and) with thicker lines and bigger fonts.} 
    Synthetic regression:
      \textit{(Left)} Test predictions compared to the ground-truth function.
      \textit{(Right)} Convergence traces for \ourmethod; green dashed line denotes the fixed \(\sigma = 0.1\) used in \textbf{Fixed Aug}.}
    \label{fig:complex_regression}
   \vspace{-1ex}
\end{figure}

We begin with a toy regression problem by generating $50$ training and $1000$ test points from 
$
y = \sin(2x) + 0.5 \cos(3x) + \varepsilon + \epsilon \sin(x) 
$
with $\varepsilon \sim \mathcal{N}(0, 0.2^2)$ and $\epsilon \sim \mathcal{N}(0, 0.15^2)$.
% Training inputs are generated by sampling 50 data points uniformly
% from \([-3, 3]\), while the test set is formed by taking $1000$ uniformly spaced points over the same interval. 
% This setup simulates a challenging regression scenario with high noise levels
% and a multi-component signal.
%
We report results in \cref{fig:complex_regression}. 
In the competing approaches, \textbf{Fixed Aug} augments data by adding Gaussian noise with fixed standard deviation \(\sigma = 0.1\).
In \textbf{Na\"ive Aug}, for each training example, we average the loss over \(K=5\) independent augmentations with \(\sigma = 0.1\). 
In \ourmethod, the augmentation shift \(\mbgamma \sim \N(\mu, \sigma^2)\) has learnable parameters and it has a prior \(\N(0, 0.2^2)\). % We allow different learning rates for model and augmentation parameters.

%We train each method for 3000 epochs, recording both training and test mean squared error (MSE).
%\cref{fig:complex_regression} (top-left panel) shows the predictions on the test data (solid lines) for each method,
%and the ground-truth function (dashed black) without noise for reference.

% \cref{tab:synthetic_complex} summarizes the final MSE:
% \begin{table}[t]
%     \centering
%     \caption{\maurizio{limit the numbers to three significant digits} Final MSE on the synthetic regression task with increased noise and a more complex underlying function.
%     \bahien{I would remove this table as the figure \cref{fig:complex_regression}.}
%     }
%     \label{tab:synthetic_complex}
%     \small
%     \begin{tabular}{lccc}
%     \toprule
%     \textbf{Method} & \textbf{Train MSE} & \textbf{Test MSE} & \textbf{Noise \(\sigma\)} \\
%     \midrule
%     No Aug & 0.02120 & 0.10176 & - \\
%     Fixed Aug & 0.03905 & 0.07412 & 0.1 \\
%     na\"ive Aug (\(K=5\)) & 0.03929 & 0.07346 & 0.1 \\
%     Learnable Aug & 0.04699 & 0.07177 & $\approx 0.1794$ (learned) \\
%     \bottomrule
%     \end{tabular}
% \end{table}

%\vspace{-0.3ex}

Although \textbf{No Aug} attains a lower training error, its test error is significantly higher due to overfitting.
Conversely, \textbf{Fixed Aug} and \textbf{Na\"ive Aug} achieve better test performance than no augmentation,
indicating that input perturbations help regularize the model.
Our \ourmethod achieves competitive test MSE.
The learned augmentation distribution widens over training
taking \(\sigma\) from $0.10$ to about $0.18$, implying that a broader range of translational perturbations is optimal
for this dataset.
This dynamic adaption shows the benefit of OPTIMA’s ability to learn the augmentation distribution, and it is theoretically justified in \cref{cor:optimal-variance}  and \cref{cor:data-drive-aug}, which state that \ourmethod tailors the augmentation distribution to the observed data. For additional ablations with different intensities on image classification dataset \cifar, see \cref{app:cifar10:svhn}.

%%\vspace{-1.8ex}

\subsection{\imagenet and \imagenetc}
\label{subsec:exp-imagenet}

%\vspace{-1.2ex}

\begin{wraptable}[7]{r}{0.4\textwidth}
    \vspace{-5mm}
    \centering
    \caption{ \imagenet and \imagenetc with non-Bayesian ResNet-50.} 
    \label{tab:ImageNet:resnet}

   \vspace{-1.5ex}
   
     \scalebox{.88}{
        \begin{tabular}{lcc}
        \toprule
        Method   & Acc (\%)  & Acc (\%) \\
           &  Clean  & Corrupted \\
        \midrule
        Mixup  & 76.1 & 40.1  \\
        \ourmethod Mixup  & \textbf{76.8}  & \textbf{41.6}      \\
        % Cutmix &        &      \\  
        % Cutmix-BA & 76.1   & 37.0               \\    
        \bottomrule
    \end{tabular}
    }
\end{wraptable}

We next evaluate the robustness of \ourmethod on \imagenet \citep{deng2009imagenet} and \imagenetc, an out-of-distribution (OOD) dataset \citep{hendrycks2019benchmarking} using a Bayesian ResNet-18 \citep{he2016deep}, where the final layer is replaced with a \texttt{BayesianLinear} module. This partially stochastic design—treating only the final layer in a Bayesian manner—is a common and efficient strategy in Bayesian deep learning \citep{harrison2024variation}. As noted by \citet{sharma2023pstoch}, full network stochasticity is often unnecessary; introducing stochasticity in the final layer can be sufficient to capture predictive uncertainty, especially with strong deterministic feature extractors \citep{kristiadi2020being}. This allows the model to represent uncertainty in class probabilities while leveraging the pretrained backbone. With \ourmethod, we optimize augmentation parameters for Mixup \citep{zhang2017mixup}, CutMix \citep{yun2019cutmix}, and AugMix \citep{hendrycks2019augmix}. Importantly, our approach is general and can also be applied to standard (non-Bayesian) networks. To illustrate this, we further evaluate \ourmethod on ResNet-50 without Bayesian treatment of its parameters. Implementation details are provided in \cref{app:imagenet:implementdetails}.

%\vspace{-0.5ex}

% We employ \MC predictions with $T = 50$ stochastic forward passes.
\cref{tab:ImageNet:mixes} summarizes the results, confirming that \ourmethod obtains better calibration and robustness for both ID and OOD. Regarding the non-Bayesian NNs, our framework also allows us to have better accuracy on the clean and corrupted data (see \cref{tab:ImageNet:resnet}), further demonstrating that \ourmethod captures variations in the data better than fixed augmentations.
More results can be found in \cref{sec:app:resnet50-nonbayesian}.
These results supports our theoretical analyses -- e.g, \cref{thm:generalization-advantage} (improved generalization on test and OOD data) and \cref{thm:posterior-shrinkage} (enhanced calibration and uncertainty quantification).

\begin{table}[ht]
\centering
\caption{\imagenet and \imagenetc with pretrained Bayesian ResNet-50 (last layer) after 10 epochs ("C"-corrupted, "m" - mean)}
\label{tab:ImageNet:mixes}

\vspace{-1.5ex}

\begin{adjustbox}{width=\textwidth}
\begin{tabular}{lccccc}
\toprule
Method & Test Acc (Clean Data) (\%) & \ECE ($\downarrow$) & mCE ($\downarrow$) (unnormalized) (C) (\%) & m\ECE ($\downarrow$) (C) & mOOD-AUROC (C)\\
\midrule       
Fixed Mixup  & 75.39 & 0.043  & 61.69 & 0.062 & 0.820 \\
\cellcolor[HTML]{edf7f0}\ourmethod Mixup  & \cellcolor[HTML]{edf7f0}74.97 & \cellcolor[HTML]{edf7f0}\textbf{0.031}  & \cellcolor[HTML]{edf7f0}\textbf{61.65} & \cellcolor[HTML]{edf7f0}\textbf{0.045} & \cellcolor[HTML]{edf7f0}\textbf{0.822} \\
Fixed Cutmix & 74.17 & 0.036  & 63.28 & 0.059 & 0.819 \\
\cellcolor[HTML]{edf7f0}\ourmethod Cutmix  & \cellcolor[HTML]{edf7f0}\textbf{74.34} & \cellcolor[HTML]{edf7f0}\textbf{0.034}  & \cellcolor[HTML]{edf7f0}63.60 & \cellcolor[HTML]{edf7f0}\textbf{0.058} & \cellcolor[HTML]{edf7f0}\textbf{0.820} \\
Fixed Augmix & 74.71 & 0.084  & 61.45 & 0.156 & 0.790 \\
\cellcolor[HTML]{edf7f0}\ourmethod Augmix & \cellcolor[HTML]{edf7f0}\textbf{75.33} & \cellcolor[HTML]{edf7f0}\textbf{0.083}  & \cellcolor[HTML]{edf7f0}\textbf{60.68} & \cellcolor[HTML]{edf7f0}\textbf{0.149} & \cellcolor[HTML]{edf7f0}\textbf{0.793} \\
\bottomrule
\end{tabular}
\end{adjustbox}
\vspace{-2ex}
\end{table}

\subsection{Computational Efficiency and Comparison with Bayesian Optimization}
\label{subsec:cifar10:bo}

% \begin{figure}[ht]
% \centering
% [PLACEHOLDER FOR FIGURE: Visualization of learned augmentations]
% \caption{Examples of augmentations sampled from the learned distribution.}
% \label{fig:learned-aug}
% \end{figure}

Our method introduces almost no additional computational cost compared to traditional data augmentation. The difference lies in our adaptive augmentation strategies, which evolve over iterations rather than remaining fixed. \ourmethod employs Monte Carlo estimates with a small sample size like one per data point per iteration and uses the reparameterization trick for efficient, low-variance gradient estimation.
To highlight its efficiency, we compare against Bayesian Optimization (BO), a strong baseline for augmentation tuning. BO requires costly black-box optimization with many full training runs per hyperparameter setting, whereas OPTIMA’s tractable ELBO jointly optimizes augmentation and model parameters within the same training loop—removing the need for separate validation runs.

% Our method adds negligible computational overhead compared to standard data augmentation. The key difference is adaptive augmentations that evolve during training rather than remaining fixed. \ourmethod uses a single Monte Carlo sample per data point per iteration and leverages the reparameterization trick for efficient, low-variance gradients.
% To demonstrate efficiency, we compare against Bayesian Optimization (BO), a strong tuning baseline. BO relies on costly black-box search with multiple full training runs per hyperparameter setting, whereas OPTIMA jointly optimizes augmentation and model parameters within one training loop, eliminating separate validation runs.

We evaluate on \cifar \citep{krizhevsky2009cifar10} using a pretrained Bayesian ResNet-18 (Bayesian last layer) to optimize augmentation parameters (mean and variance). BO is run for 25 trials of 15 epochs, followed by 50 epochs of training with the optimized parameters, while \ourmethod is trained directly for 50 epochs. For augmentation, we use Mixup and learn the parameter $\alpha$. We also assess performance on \cifar-C \citep{hendrycks2019benchmarking} as OOD data. As shown in \cref{tab:cifar10:bo_main}, \ourmethod achieves higher test accuracy on clean data (with a slight calibration trade-off) and substantially better accuracy, ECE, and AUROC on OOD data, all in far less time than BO—demonstrating improved calibration and robustness at much lower cost.
\begin{table}[ht]
\centering
\caption{Comparison between Bayesian optimization and \ourmethod on \cifar}
\label{tab:cifar10:bo_main}

\vspace{-1.5ex}

\scalebox{.95}{
    \begin{adjustbox}{width=\textwidth}
    \begin{tabular}{lcccccc}
    \toprule
    Method & Test Acc (\%) & \ECE (\%) & mAccuracy (C) & m\ECE (C) & OOD AUROC & Time\\
    \midrule
    Bayesian Optimization & 93.43 & 0.010 & 72.44 & 0.127 & 0.652 & $\sim 4 \times T$ \\
    \ourmethod & \textbf{95.03} & 0.047 & \textbf{78.52} & \textbf{0.076} & \textbf{0.680}  & $T$\\
    \bottomrule
    \end{tabular}
    \end{adjustbox}
}

\vspace{-2.5ex}

\end{table}

\subsection{OPTIMA on Discrete NLP Augmentations: SST-5 Case Study}
\label{subsec:sst5}

To show that OPTIMA is not restricted to continuous or geometric
transformations used in computer vision, we additionally evaluate it on a NLP classification task where augmentations are inherently
\emph{discrete}. We use the SST-5 benchmark (\citealt{socher2013recursive}),
a fine-grained 5-class sentiment dataset, and fine-tune a DistilBERT model
(\citealt{sanh2019distilbert}) for 5 epochs on the full training split.

\paragraph{Discrete augmentation family.}
% We consider \emph{token dropout}, a stochastic masking transformation widely
% used in NLP regularization. For an input sequence $x=(x_1,\dots,x_L)$ and a
% dropout probability $p_{\mathrm{drop}}$, the augmentation samples a Bernoulli
% mask $\gamma_t \sim \mathrm{Bernoulli}(1 - p_{\mathrm{drop}})$ and replaces
% $x_t$ with \texttt{[MASK]} whenever $\gamma_t = 0$. 
% This produces a discrete latent transformation variable~$\gamma$. 
% Although the transformation is non-differentiable, OPTIMA can still optimize 
% the dropout probability by using a score-function (REINFORCE) gradient, as 
% predicted by our general formulation in Section~\ref{sec:method}.

% We consider \emph{token dropout}, a stochastic masking transformation widely used in NLP regularization. For an input sequence $x=(x_1,\dots,x_L)$ and dropout probability $p_{\mathrm{drop}}$, we sample a Bernoulli mask $\gamma_t \sim \mathrm{Bernoulli}(1 - p_{\mathrm{drop}})$ and replace $x_t$ with \texttt{[MASK]} whenever $\gamma_t = 0$. This yields a discrete latent transformation variable~$\gamma$. Although non-differentiable, OPTIMA can still optimize $p_{\mathrm{drop}}$ using a score-function (REINFORCE) gradient, as predicted by our general formulation in Section~\ref{sec:method}.

We consider \emph{token dropout}, a stochastic masking transformation used in NLP regularization. For input $x=(x_1,\dots,x_L)$ with dropout rate $p_{\mathrm{drop}}$, we sample $\gamma_t \sim \mathrm{Bernoulli}(1 - p_{\mathrm{drop}})$ and replace $x_t$ with \texttt{[MASK]} if $\gamma_t=0$, yielding a discrete latent variable~$\gamma$. Although non-differentiable, OPTIMA optimizes $p_{\mathrm{drop}}$ via a score-function (REINFORCE) gradient, consistent with Section~\ref{sec:method}.

\vspace{-1.0ex}

\paragraph{Experimental setup.}
We evaluate OPTIMA on discrete token-dropout augmentation. The augmentation
uses a dropout probability $p_{\mathrm{drop}} \in [0,p_{\max}]$ parameterized
as $p_{\mathrm{drop}} = p_{\max}\,\sigma(s)$, where $s$ is a learnable scalar.
To encode prior preferences for dropout, we place a
Gaussian prior directly on $p_{\mathrm{drop}}$, and OPTIMA jointly learns $s$
and the model parameters via our \ELBO. We compare OPTIMA
against: (i) \emph{No Aug}; (ii) \emph{Fixed Aug}, which uses the same initial
dropout as OPTIMA; (iii) \emph{Fixed Aug (Matched)}, where $p_{\mathrm{drop}}$
is set equal to the value learned by OPTIMA; and (iv) \emph{BO-Fixed}, which
selects $p_{\mathrm{drop}}$ through a validation-based hyperparameter
search. More details are provided in \cref{app:sst5}.

\begin{table}[ht]
\centering
\caption{SST-5 results for discrete token-dropout augmentation averaged over 5 different seeds.}
\label{tab:sst5_optima}

\vspace{-1ex}

\scalebox{0.95}{
\begin{tabular}{lccc}
\toprule
\textbf{Method} & \textbf{Accuracy} & \textbf{NLL} & \textbf{ECE} \\
\midrule
No Aug & $0.516 \pm 0.003$ & $1.240 \pm 0.010$ & $0.190 \pm 0.004$ \\
Fixed $p_{\text{drop}} = 0.04$ & $0.522 \pm 0.003$ & $1.180 \pm 0.006$ & $0.154 \pm 0.006$ \\
Fixed $p_{\text{drop}} = 0.0625$ & $0.516 \pm 0.006$ & $1.162 \pm 0.007$ & $0.143 \pm 0.007$ \\
\ourmethod{} with $\mu = 0.1$ ($p_{\text{learned}}=  0.0625$) & \textbf{0.524} $\pm$ \textbf{0.003} & \textbf{1.161} $\pm$ \textbf{0.007} & \textbf{0.142} $\pm$ \textbf{0.006} \\
BO-Fixed $p_{\text{drop}} = 0.3$ & $0.521 \pm 0.004$ & $1.086 \pm 0.006$ & \textbf{0.043} $\pm$ \textbf{0.004} \\
\ourmethod{} with $\mu = 0.3$ ($p_{\text{learned}} = 0.3$) & \textbf{0.524} $\pm$ \textbf{0.004} & \textbf{1.086} $\pm$ \textbf{0.005} & \textbf{0.046} $\pm$ \textbf{0.002} \\
\bottomrule
\end{tabular}
}
\end{table}

\vspace{-1.5ex}

% \paragraph{Results.}
% Table~\ref{tab:sst5_optima} reports accuracy, NLL, and ECE. Although accuracy differences are small—consistent with prior findings that SST-5 is difficult for smaller Transformers—OPTIMA achieves \emph{lower NLL} and markedly better calibration than fixed-augmentation baselines. Crucially, OPTIMA also improves over the matched-strength dropout baseline, showing that its gains stem from optimizing the \emph{marginal likelihood} rather than simply selecting a favorable dropout rate.

% These results demonstrate that OPTIMA extends naturally to discrete augmentations and that its theoretical advantages (Sections~\ref{sec:generalization}--\ref{sec:invariance}) persist beyond vision tasks.

% \paragraph{Results.}
% Table~\ref{tab:sst5_optima} shows that accuracy differences are expectedly small for SST-5, but OPTIMA consistently achieves \emph{lower NLL} and better calibration than fixed-augmentation baselines.
% Importantly, OPTIMA also matches the BO-tuned baseline—despite BO requiring a full hyperparameter search over multiple training runs (\emph{it took around 8$
% \times$ times more}), whereas OPTIMA learns $p_{\mathrm{drop}}$ in a \emph{single} training run.
% This highlights that the gains arise from optimizing the \emph{marginal likelihood} rather than simply selecting a favorable dropout rate. These results confirm that OPTIMA naturally extends to discrete augmentation spaces and that its theoretical advantages (Sections~\ref{sec:generalization}--\ref{sec:invariance}) persist beyond vision tasks.

\paragraph{Results.}
Table~\ref{tab:sst5_optima} shows that accuracy differences on SST-5 are small, but OPTIMA consistently achieves lower \emph{NLL} and improved calibration over fixed-augmentation baselines. Notably, it matches the BO-tuned baseline—despite BO requiring a full hyperparameter search across multiple runs (approximately $8\times$ more compute)—while OPTIMA learns $p_{\mathrm{drop}}$ in a single training run. This indicates that the gains stem from marginal likelihood optimization rather than dropout tuning. These findings confirm that OPTIMA extends naturally to discrete augmentation spaces and that its theoretical advantages (Sections~\ref{sec:generalization}--\ref{sec:invariance}) generalize beyond vision tasks.

%------------------------------------------------------------------------------%
\section{Discussion and Conclusion}
\label{sec:discussion}

%%\vspace{-1.5ex}

\vspace{-1.2ex}

We presented a theoretical and methodological framework for optimizing \da taking inspiration from Bayesian principles, which allow us to cast this problem as  model selection. 
We derived a variational objective to learn optimal \da strategies from data in a practical way.
We also provided extensive theoretical insights on the advantages of our proposed data-driven approach to \da compared to alternatives, revealing improved generalization through PAC-Bayes bounds, enhanced invariance via higher-order regularization, and better calibration through marginalization.
Empirical results confirm these theoretical benefits, showing consistent improvements in calibration and predictive performance across various tasks. 
We believe that \ourmethod is a key step toward robust and well-calibrated models capable of assisting decision-making in applications where this is of critical importance.

% We introduced a principled framework for \emph{Bayesian-optimized \da}, where augmentation parameters are treated as latent variables within a variational Bayesian objective. Our theoretical results show that this integration:
% \begin{itemize}
%     \item \emph{Improves generalization}, supported by new PAC-Bayes bounds specialized to augmented likelihoods.
%     \item \emph{Enforces invariance} via both Jacobian (first-order) and Hessian (second-order) regularization, flattening the loss landscape and enhancing robustness.
%     \item \emph{Avoids overcounting} by marginalizing out augmentation parameters, yielding better calibration.
%     \item \emph{Learns augmentations} under a Bayesian perspective, discovering transformations that best help posterior inference.
% \end{itemize}

% Our empirical results on synthetic regression and \cifar classification confirm these theoretical advantages, showing consistent improvements in test performance, calibration, and out-of-distribution robustness compared to fixed or no augmentation.

%\vspace{-2.1ex}

\vspace{-1.8ex}

\paragraph{Limitations and future work.}

% While \ourmethod offers significant advantages, it has some limitations that suggest directions for future work.
% Our current implementation focuses on relatively simple transformations and mostly on computer vision application; extending to more complex, composition-based transformations and other data modalities would broaden the applicability of \ourmethod.
% Also, our theoretical analysis could be extended to provide tighter bounds and more precise characterizations of the benefits of \ourmethod.

While \ourmethod offers significant advantages, it has some limitations that suggest directions for future work.
Although our main experiments focus on computer vision, \ourmethod itself is not tied to continuous or geometric transformations. Our additional evaluation on a natural language task (SST-5) demonstrates that OPTIMA can also handle discrete, non-geometric transformations by optimizing a latent augmentation distribution in text space. Nevertheless, a broader exploration of more expressive or compositional transformations in NLP, time series, or multimodal settings remains an important next step.
In addition, our theoretical analysis could be strengthened by developing tighter PAC-Bayes bounds and more refined characterizations of the benefits introduced by Bayesian marginalization over augmentation parameters.

\bibliography{refs.bib}
\bibliographystyle{iclr2026_conference}

\newpage
\appendix
% \section{Appendix}

{\bf \Large{Appendix}}

\begin{spacing}{0.2}
    \addtocontents{toc}{\protect\setcounter{tocdepth}{2}} % Set level of depth to 3
    \renewcommand{\contentsname}{\textbf{Table of Contents}\vskip3pt\hrule}
    \tableofcontents
    \vskip5pt\hrule
\end{spacing}

\section{Derivation of the Augmented Evidence Lower Bound}
\label{subsec:augmented-elbo}

For variational inference, we introduce a variational distribution \(q(\mbtheta, \mbphi) = q(\mbtheta) q(\mbphi)\) to approximate the posterior \(p(\mbtheta, \mbphi \g \mathcal{D})\). The standard \ELBO is a lower bound on the log marginal likelihood \( \log p(\mathcal{D}) = \log \iiint p(\mathcal{D}, \mbtheta, \mbphi, \mbgamma) \, d\mbtheta \, d\mbphi \, d\mbgamma \).
% \begin{align}
% \log p(\mathcal{D}) = \log \int \int \int p(\mathcal{D},\mbtheta,\mbphi, \mbgamma) \, d\mbtheta \, d\mbphi \, d\mbgamma.    
% \end{align}
Using Jensen’s inequality with \(q(\mbtheta,\mbphi)\), we have:
\begin{align}
    \log p(\mathcal{D}) \geq \underbrace{\mathbb{E}_{q(\mbtheta, \mbphi)} \left[ \log \frac{p(\mathcal{D}, \mbtheta, \mbphi)}{q(\mbtheta, \mbphi)} \right]}_{\mathrm{ELBO}(q)}, \quad \text{where \(p(\mathcal{D}, \mbtheta, \mbphi) = \int p(\mathcal{D}, \mbtheta, \mbphi, \mbgamma) \, d\mbgamma\)}. \label{eq:elbo}
\end{align}

% The likelihood term is:

% \[
% p(\mathcal{D} \g\mbtheta,\mbphi) = \mathbb{E}_{p(\mbgamma \g\mbphi)} \left[ \prod_{i=1}^N p(y_i \g T_\mbgamma(x_i),\mbtheta) \right].
% \]

Applying Jensen’s inequality further to the log of the likelihood term, $p(\mathcal{D} \g \mbtheta, \mbphi)$:

\begin{align}
\hspace{-1.7ex} \log p(\mathcal{D} \g \mbtheta, \mbphi) = \log \mathbb{E}_{p(\mbgamma \g \mbphi)} \left[ \prod_{i=1}^N p(\mby_i \g T_\mbgamma(\mbx_i), \mbtheta) \right] \geq \mathbb{E}_{p(\mbgamma \g\mbphi)} \left[ \sum_{i=1}^N \log p(\mby_i \g T_\mbgamma(\mbx_i),\mbtheta) \right].
\end{align}

Substituting this into the \ELBO in \cref{eq:elbo}, with $p(\mathcal{D},   \mbtheta, \mbphi) = p(\mathcal{D} \g \mbtheta, \mbphi) p (\mbtheta) p(\mbphi)$, we can obtain the augmented \ELBO as follows:

% \[
% \mathrm{ELBO}(q) = \mathbb{E}_{q(\mbtheta,\mbphi)} \left[ \log p(\mathcal{D} \g\mbtheta,\mbphi) + \log p(\mbtheta) + \log p(\mbphi) - \log q(\mbtheta) - \log q(\mbphi) \right].
% \]

% Using the lower bound for \(\log p(\mathcal{D} \g\mbtheta,\mbphi)\):

% \begin{align}
%     \hspace{-3ex}  \mathrm{ELBO}&(q) \geq \mathbb{E}_{q(\mbtheta,\mbphi)} \left[ \mathbb{E}_{p(\mbgamma \g\mbphi)} \left[ \sum_{i=1}^N \log p(\mby_i \g T_\mbgamma(\mbx_i),\mbtheta) \right] + \log \frac{p(\mbtheta) p(\mbphi)}{q(\mbtheta) q(\mbphi)} \right] \\
%     &=  \underbrace{ {\textcolor{BlueViolet}{\mathbb{E}_{q(\mbtheta)}  \mathbb{E}_{q(\mbphi)} \mathbb{E}_{p(\mbgamma \g\mbphi)} \left[ \sum_{i=1}^N \log p(\mby_i \g T_\mbgamma(\mbx_i),\mbtheta) \right]}} - {\textcolor{Bittersweet}{\text{KL}(q(\mbtheta) \| p(\mbtheta))}} - {\textcolor{purple}{\text{KL}(q(\mbphi) \| p(\mbphi))}}}_{\mathrm{ELBO}_{\text{aug}}(q_\mbtheta, q_\mbphi)}.
% \end{align}

\begin{align}
    \hspace{-5ex} &\mathrm{ELBO}(q) \geq \mathbb{E}_{q(\mbtheta,\mbphi)} \left[ \mathbb{E}_{p(\mbgamma \g\mbphi)} \left[ \sum_{i=1}^N \log p(\mby_i \g T_\mbgamma(\mbx_i),\mbtheta) \right] + \log \frac{p(\mbtheta) p(\mbphi)}{q(\mbtheta) q(\mbphi)} \right] \\
    &=  \underbrace{ {\textcolor{BlueViolet}{\mathbb{E}_{q(\mbtheta)}  \mathbb{E}_{q(\mbphi)} \mathbb{E}_{p(\mbgamma \g\mbphi)} \left[ \sum_{i=1}^N \log p(\mby_i \g T_\mbgamma(\mbx_i),\mbtheta) \right]}}}_{\text{data fit}} - \underbrace{{\textcolor{Bittersweet}{\text{KL}(q(\mbtheta) \| p(\mbtheta))}}}_{\text{parameter prior}} - \underbrace{{\textcolor{purple}{\text{KL}(q(\mbphi) \| p(\mbphi))}}}_{\text{augmentation prior}}. \label{eq:aug_elbo_def:appendix}
\end{align}

% Expanding with \(q(\mbtheta,\mbphi) = q(\mbtheta) q(\mbphi)\):

% \[
% = \mathbb{E}_{q(\mbtheta)} \mathbb{E}_{q(\mbphi)} \mathbb{E}_{p(\mbgamma \g\mbphi)} \left[ \sum_{i=1}^N \log p(y_i \g T_\mbgamma(x_i),\mbtheta) \right] - \text{KL}(q(\mbtheta) \| p(\mbtheta)) - \text{KL}(q(\mbphi) \| p(\mbphi)).
% \]

% Thus, the augmented ELBO is:

% \begin{equation}
% \mathrm{ELBO}_{\text{aug}}(q_\mbtheta, q_\mbphi) = \mathbb{E}_{q(\mbtheta)} \mathbb{E}_{q(\mbphi)} \mathbb{E}_{p(\mbgamma \g\mbphi)} \left[ \sum_{i=1}^N \log p(y_i \g T_\mbgamma(x_i),\mbtheta) \right] - \text{KL}(q(\mbtheta) \| p(\mbtheta)) - \text{KL}(q(\mbphi) \| p(\mbphi)).
% \label{eq:augmented-elbo}
% \end{equation}

% This augmented ELBO consists of three terms:
% - The first term encourages the model to fit the data, averaging over the variational distributions of \(\mbtheta\) and \(\mbphi\), and the augmentation distribution \(p(\mbgamma \g\mbphi)\).
% - The second term, \(\text{KL}(q(\mbtheta) \| p(\mbtheta))\), regularizes the model parameters.
% - The third term, \(\text{KL}(q(\mbphi) \| p(\mbphi))\), regularizes the augmentation parameters, ensuring they align with the prior \(p(\mbphi)\).

The augmented \ELBO consists of three terms: a \textcolor{BlueViolet} {data-fitting term} that averages over the variational distributions \(q(\mbtheta)\), \(q(\mbphi)\), and the augmentation distribution \(p(\mbgamma \mid \mbphi)\); a \textcolor{Bittersweet}{regularization term \(\text{KL}(q(\mbtheta) \| p(\mbtheta))\)} that penalizes divergence from the prior over model parameters; and another \textcolor{purple}{regularization term \(\text{KL}(q(\mbphi) \| p(\mbphi))\)} that aligns the augmentation parameters with their prior.

% \paragraph{Appendix}
\section{Detailed Proofs}
\label{app:proofs}

This section provides expanded proofs for the theoretical results presented in the main paper.

\subsection{Proof of \cref{prop:jensen-gap} (Jensen Gap Bound)} \label{sec:proof_prop_jensen-gap}

% \begin{proof}
% Let \(f(\mbgamma) = \log p(y|T_\mbgamma(x),\mbtheta)\). Suppose \(f\) is \(L\)-Lipschitz in \(\mbgamma\). Denote by \(\mu = \E[\mbgamma]\). For sub-Gaussian \(\mbgamma\) with variance proxy \(\sigma^2\), we use standard mgf bounds:
% \begin{align}
% \log \E_{\mbgamma}[e^{f(\mbgamma)}] &= \log \E_{\mbgamma}\Bigl[e^{f(\mu)+(f(\mbgamma)-f(\mu))}\Bigr] \\
% &= f(\mu) + \log\E_{\mbgamma}\Bigl[e^{f(\mbgamma)-f(\mu)}\Bigr].
% \end{align}

% Since \(\|f(\mbgamma)-f(\mu)\| \leq L \|\mbgamma-\mu\|\), the sub-Gaussian property implies:
% \begin{align}
% \E_{\mbgamma}[e^{f(\mbgamma)-f(\mu)}] \leq \E_{\mbgamma}[e^{L\|\mbgamma-\mu\|}] \leq e^{\frac{L^2\sigma^2}{2}}.
% \end{align}

% Thus:
% \begin{align}
% \log\E_{\mbgamma}[e^{f(\mbgamma)}] \leq f(\mu) + \frac{L^2\sigma^2}{2}.
% \end{align}

% Meanwhile, \(\E_{\mbgamma}[f(\mbgamma)] = f(\mu)\) if \(\E_{\mbgamma}[\mbgamma-\mu] = 0\). Hence:
% \begin{align}
% \text{Gap} &= \log \E_{\mbgamma}[e^{f(\mbgamma)}] - \E_{\mbgamma}[f(\mbgamma)] \\
% &\leq \frac{L^2\sigma^2}{2}.
% \end{align}

% For tightness, when \(f(\mbgamma)\) is approximately linear in the high-probability region of \(p(\mbgamma|\mbphi)\), the bound approaches equality.
% \end{proof}

\begin{proof}
Let \(f(\mbgamma) = \log p(\mby \g T_\mbgamma(\mbx),\mbtheta)\). For sub-Gaussian \(\mbgamma\) with mean \(\mu = \E[\mbgamma]\) and variance proxy \(\sigma^2\), we use standard moment generating function bounds:
\begin{align}
\log \E_{\mbgamma}[e^{f(\mbgamma)}] &= \log \E_{\mbgamma}\Bigl[e^{f(\mu)+(f(\mbgamma)-f(\mu))}\Bigr] \\
&= f(\mu) + \log\E_{\mbgamma}\Bigl[e^{f(\mbgamma)-f(\mu)}\Bigr].
\end{align}

Since \(f\) is \(L\)-Lipschitz, we have \(|f(\mbgamma)-f(\mu)| \leq L \|\mbgamma-\mu\|\). Using the sub-Gaussian property:
\begin{align}
\E_{\mbgamma}[e^{f(\mbgamma)-f(\mu)}] \leq \E_{\mbgamma}[e^{L\|\mbgamma-\mu\|}] \leq e^{\frac{L^2\sigma^2}{2}}.
\end{align}

Therefore:
\begin{align}
\log\E_{\mbgamma}[e^{f(\mbgamma)}] \leq f(\mu) + \frac{L^2\sigma^2}{2}.
\end{align}

Since \(\E_{\mbgamma}[f(\mbgamma)] = f(\mu)\) when \(\E_{\mbgamma}[\mbgamma-\mu] = 0\), the gap is:
\begin{align}
\text{Gap} &= \log \E_{\mbgamma}[e^{f(\mbgamma)}] - \E_{\mbgamma}[f(\mbgamma)] \\
&\leq \frac{L^2\sigma^2}{2}.
\end{align}

For tightness, when \(f(\mbgamma)\) is approximately linear in the high-probability region of \(p(\mbgamma|\mbphi)\), the bound approaches equality.
\end{proof}

% \subsection{Proof of Theorem~\cref{thm:pac-bayes} (PAC-Bayes Under Augmentation)}

% \begin{proof}
% Define the loss function as:
% \[
% \ell(\mbtheta,\mbphi, (x, y)) = -\log \mathbb{E}_{p(\mbgamma \mid\mbphi)} [p(y \mid T_\mbgamma(x),\mbtheta)].
% \]

% For an i.i.d. dataset \(\mathcal{D} = \{(x_i, y_i)\}_{i=1}^N\), the empirical risk is:
% \[
% \hat{R}(\mbtheta,\mbphi) = \frac{1}{N} \sum_{i=1}^N \ell(\mbtheta,\mbphi, (x_i, y_i)).
% \]

% Since \(\{(x_i, y_i)\}_{i=1}^N\) are i.i.d., and the expectation over \(\mbgamma \sim p(\mbgamma \mid\mbphi)\) is computed independently for each sample, the terms \(\ell(\mbtheta,\mbphi, (x_i, y_i))\) are independent for fixed \(\mbtheta,\mbphi\). Applying the standard PAC-Bayes theorem over the joint space \((\mbtheta,\mbphi)\):
% \[
% \mathbb{E}_{q(\mbtheta,\mbphi)} [R(\mbtheta,\mbphi)] \leq \mathbb{E}_{q(\mbtheta,\mbphi)} [\hat{R}(\mbtheta,\mbphi)] + \sqrt{\frac{\text{KL}(q(\mbtheta,\mbphi) \| p(\mbtheta,\mbphi)) + \log \frac{2 \sqrt{N}}{\delta}}{2N}}.
% \]

% This completes the proof.
% \end{proof}

\subsection{Proof of \cref{thm:pac-bayes} (PAC-Bayes Under Augmentation)} \label{sec:proof_thm_pac_bayes}

\begin{proof}
Define the loss function as:
\begin{align}
\ell(\mbtheta,\mbphi, (x, y)) = -\log \mathbb{E}_{p(\mbgamma \mid\mbphi)} [p(y \mid T_\mbgamma(x),\mbtheta)].
\end{align}

The empirical risk is:
\begin{align}
\hat{R}(\mbtheta,\mbphi) = \frac{1}{N} \sum_{i=1}^N \ell(\mbtheta,\mbphi, (\mbx_i, \mby_i)).
\end{align}

Since \(\{(\mbx_i, \mby_i)\}_{i=1}^N\) are i.i.d., and the expectation over \(\mbgamma \sim p(\mbgamma \mid\mbphi)\) is computed independently for each sample, the terms \(\ell(\mbtheta,\mbphi, (\mbx_i, \mby_i))\) are independent for fixed \(\mbtheta,\mbphi\). Applying the standard PAC-Bayes theorem over the joint space \((\mbtheta,\mbphi)\) \citep{mcallester1999pac,catoni2007pac,alquier2024user}: 
\begin{align}
\mathbb{E}_{q(\mbtheta,\mbphi)} [R(\mbtheta,\mbphi)] \leq \mathbb{E}_{q(\mbtheta,\mbphi)} [\hat{R}(\mbtheta,\mbphi)] + \sqrt{\frac{\text{KL}(q(\mbtheta,\mbphi) \| p(\mbtheta,\mbphi)) + \log \frac{2 \sqrt{N}}{\delta}}{2N}}.
\end{align}

This completes the proof.
\end{proof}

\subsection{Proof of \cref{thm:generalization-advantage} (Generalization Advantange of Bayesian-Optimized Augmentation) } \label{sec:proof_thm_generalization_advantage}

\begin{proof}
\textbf{Step 1: PAC-Bayes Bounds}

For \ourmethod, the PAC-Bayes bound is:

\begin{align}
\mathbb{E}_{q(\mbtheta,\mbphi)} [R(\mbtheta,\mbphi)] \leq \mathbb{E}_{q(\mbtheta,\mbphi)} [\hat{R}(\mbtheta,\mbphi)] + \sqrt{\frac{\text{KL}(q(\mbtheta,\mbphi) \| p(\mbtheta,\mbphi)) + \log \frac{2 \sqrt{N}}{\delta}}{2N}}.
\end{align}

For na\"ive augmentation, the bound is:

\begin{align}
\mathbb{E}_{q(\mbtheta)} [R(\mbtheta)] \leq \mathbb{E}_{q(\mbtheta)} [\hat{R}_{\text{na\"ive}}(\mbtheta)] + \sqrt{\frac{\text{KL}(q(\mbtheta) \| p(\mbtheta)) + \log \frac{2 \sqrt{N}}{\delta}}{2N}}.
\end{align}

\textbf{Step 2: Relationship Between Empirical Risks}

By Jensen's inequality, for each data point \((\mbx_i, \mby_i)\):

\begin{align}
\log \mathbb{E}_{p(\mbgamma \mid\mbphi)} p(\mby_i \mid T_\mbgamma(\mbx_i),\mbtheta) \geq \mathbb{E}_{p(\mbgamma \mid\mbphi)} \log p(\mby_i \mid T_\mbgamma(\mbx_i),\mbtheta).
\end{align}

Thus,

\begin{align}
-\log \mathbb{E}_{p(\mbgamma \mid\mbphi)} p(\mby_i \mid T_\mbgamma(\mbx_i),\mbtheta) \leq -\mathbb{E}_{p(\mbgamma \mid\mbphi)} \log p(\mby_i \mid T_\mbgamma(\mbx_i),\mbtheta).
\end{align}

For large \(K\), the na\"ive empirical risk approximates:

\begin{align}
\hat{R}_{\text{na\"ive}}(\mbtheta) \approx -\frac{1}{N} \sum_{i=1}^N \mathbb{E}_{p(\mbgamma \mid\mbphi)} \log p(\mby_i \mid T_\mbgamma(\mbx_i),\mbtheta).
\end{align}

Therefore,

\begin{align}
\hat{R}(\mbtheta,\mbphi) = -\frac{1}{N} \sum_{i=1}^N \log \mathbb{E}_{p(\mbgamma \mid\mbphi)} p(\mby_i \mid T_\mbgamma(\mbx_i),\mbtheta) \leq \hat{R}_{\text{na\"ive}}(\mbtheta),
\end{align}

with equality only if \(\Delta_{\mbphi}(\mbx_i, \mby_i) = 0\) for all \(i\), i.e., when \(p(\mby_i \mid T_\mbgamma(\mbx_i),\mbtheta)\) is constant across \(\mbgamma\).

\textbf{Step 3: Bound Comparison}

Assuming \(\text{KL}(q(\mbtheta,\mbphi) \| p(\mbtheta,\mbphi)) \approx \text{KL}(q(\mbtheta) \| p(\mbtheta))\), the difference in bounds is driven by the empirical risks:

\begin{align}
\mathbb{E}_{q(\mbtheta,\mbphi)} [R(\mbtheta,\mbphi)] \leq \mathbb{E}_{q(\mbtheta,\mbphi)} [\hat{R}(\mbtheta,\mbphi)] + \text{complexity term},
\end{align}

\begin{align}
\mathbb{E}_{q(\mbtheta)} [R(\mbtheta)] \leq \mathbb{E}_{q(\mbtheta)} [\hat{R}_{\text{na\"ive}}(\mbtheta)] + \text{complexity term}.
\end{align}

Since \(\hat{R}(\mbtheta,\mbphi) \leq \hat{R}_{\text{na\"ive}}(\mbtheta)\), and the complexity terms are similar, our bound is tighter. Specifically,

\begin{align}
\mathbb{E}_{q(\mbtheta,\mbphi)} [\hat{R}(\mbtheta,\mbphi)] = \mathbb{E}_{q(\mbtheta,\mbphi)} [\hat{R}_{\text{na\"ive}}(\mbtheta)] - \mathbb{E}_{q(\mbtheta,\mbphi)} \left[ \frac{1}{N} \sum_{i=1}^N \Delta_{\mbphi}(\mbx_i, \mby_i) \right],
\end{align}

leading to:

\begin{align}
\mathbb{E}_{q(\mbtheta,\mbphi)} [R(\mbtheta,\mbphi)] \leq \mathbb{E}_{q(\mbtheta)} [\hat{R}_{\text{na\"ive}}(\mbtheta)] - \Delta + \text{complexity term},
\end{align}

where \(\Delta = \mathbb{E}_{q(\mbtheta,\mbphi)} \left[ \frac{1}{N} \sum_{i=1}^N \Delta_{\mbphi}(\mbx_i, \mby_i) \right] \geq 0\).

Thus, \ourmethod’s bound is lower by \(\Delta\), proving better generalization. When \(p(\mby_i \mid T_\mbgamma(\mbx_i),\mbtheta)\) varies across \(\mbgamma\), \(\Delta > 0\), making our bound strictly tighter.
\end{proof}

\subsection{Proof of \cref{thm:invariance} (Higher-order Invariance)} \label{sec:proof_thm_invariance}

\begin{proof}
Using a second-order Taylor expansion of \( f_\mbtheta \) around \( x \):
\begin{align}
f_\mbtheta(T_\mbgamma(\mbx)) &= f_\mbtheta(\mbx) + J_f(\mbx) \delta + \frac{1}{2} \delta^T \nabla^2 f_\mbtheta(\mbx) \delta + \mathcal{O}(\|\delta\|^3), \\
f_\mbtheta(T_\mbgamma(\mbx)) - f_\mbtheta(\mbx) &= J_f(\mbx) \delta + \frac{1}{2} \delta^T \nabla^2 f_\mbtheta(\mbx) \delta + \mathcal{O}(\|\delta\|^3).
\end{align}

Squaring this difference and taking the expectation over \( p(\mbgamma|\mbphi) \):
\begin{align}
\mathbb{E}_{p(\mbgamma|\mbphi)} \left[ \| f_\mbtheta(T_\mbgamma(\mbx)) - f_\mbtheta(\mbx) \|^2 \right] &= \mathbb{E}_{p(\mbgamma|\mbphi)} \left[ \| J_f(\mbx) \delta \|^2 \right] \\
&\quad + \mathbb{E}_{p(\mbgamma|\mbphi)} \left[ \left( \frac{1}{2} \delta^T \nabla^2 f_\mbtheta(\mbx) \delta \right)^2 \right] \\
&\quad + \mathbb{E}_{p(\mbgamma|\mbphi)} \left[ 2 \left( J_f(\mbx) \delta \right)^T \left( \frac{1}{2} \delta^T \nabla^2 f_\mbtheta(\mbx) \delta \right) \right] \\
&\quad + \mathcal{O}(\|\delta\|^3).
\end{align}

The cross-term \( \mathbb{E}_{p(\mbgamma|\mbphi)} \left[ \left( J_f(\mbx) \delta \right)^T \left( \delta^T \nabla^2 f_\mbtheta(\mbx) \delta \right) \right] \) involves odd powers of \( \delta \), which vanish since \( \mathbb{E}[\delta] = 0 \). Thus:
\begin{align}
\mathbb{E}_{p(\mbgamma|\mbphi)} \left[ \| f_\mbtheta(T_\mbgamma(\mbx)) - f_\mbtheta(\mbx) \|^2 \right] &= \mathbb{E}_{p(\mbgamma|\mbphi)} \left[ \delta^T J_f(\mbx)^T J_f(\mbx) \delta \right] \\
&\quad + \frac{1}{4} \mathbb{E}_{p(\mbgamma|\mbphi)} \left[ \left( \delta^T \nabla^2 f_\mbtheta(\mbx) \delta \right)^2 \right] + \mathcal{O}(\|\delta\|^3).
\end{align}

Using properties of quadratic forms for zero-mean random variables:
\begin{align}
\mathbb{E}_{p(\mbgamma|\mbphi)} \left[ \delta^T J_f(\mbx)^T J_f(\mbx) \delta \right] &= \mathrm{Tr} \left( J_f(\mbx)^T J_f(\mbx) \Sigma_\mbphi \right), \\
\mathbb{E}_{p(\mbgamma|\mbphi)} \left[ \left( \delta^T \nabla^2 f_\mbtheta(\mbx) \delta \right)^2 \right] &= \mathrm{Tr} \left( \nabla^2 f_\mbtheta(\mbx)^T \nabla^2 f_\mbtheta(\mbx) \Sigma_\mbphi \right) + 2 \mathrm{Tr} \left( (\nabla^2 f_\mbtheta(\mbx) \Sigma_\mbphi)^2 \right).
\end{align}

For small perturbations, the dominant term is \( \mathrm{Tr} \left( \nabla^2 f_\mbtheta(\mbx)^T \nabla^2 f_\mbtheta(\mbx) \Sigma_\mbphi \right) \), and higher moments contribute to \( \mathcal{O}(\|\delta\|^3) \). Therefore:
\begin{align}
\mathbb{E}_{p(\mbgamma|\mbphi)} \left[ \| f_\mbtheta(T_\mbgamma(\mbx)) - f_\mbtheta(\mbx) \|^2 \right] &\approx \mathrm{Tr} \left( J_f(\mbx)^T J_f(\mbx) \Sigma_\mbphi \right) + \frac{1}{4} \mathrm{Tr} \left( \nabla^2 f_\mbtheta(\mbx)^T \nabla^2 f_\mbtheta(\mbx) \Sigma_\mbphi \right).
\end{align}

This approximation holds for small \( \|\delta\| \), completing the proof.
\end{proof}

\subsection{Proof of \cref{thm:posterior-shrinkage} (Posterior Shrinkage under na\"ive Augmentation)} \label{sec:proof_thm_posterior_shrinkage}

\begin{proof}
Under a locally Gaussian approximation with full-rank covariance, the posterior covariance is approximately the inverse of the Hessian of the negative log posterior at the MAP estimate. For the true posterior, marginalizing over \(\mbphi\):
\[
p(\mathcal{D} \g\mbtheta) = \int p(\mathcal{D} \g\mbtheta,\mbphi) p(\mbphi) \, d\mbphi,
\]
\begin{align}
    \Sigma_{\text{true}}^{-1} \approx -\nabla^2 \log p_{\text{true}}(\mbtheta \g \cD)|_{\mbtheta=\hat{\mbtheta}} &= -\nabla^2 \log p(\mbtheta) \\
    &- \sum_{i=1}^N \nabla^2 \log \left( \int \E_{p(\mbgamma|\mbphi)}[p(\mby_i \g T_\mbgamma(\mbx_i),\mbtheta)] p(\mbphi) \, d\mbphi \right) \Big|_{\mbtheta=\hat{\mbtheta}}. \nonumber
\end{align}

For the na\"ive posterior:
\begin{align}
\Sigma_{\text{na\"ive}}^{-1} \approx -\nabla^2 \log p_{\text{na\"ive}}(\mbtheta \g \cD)|_{\mbtheta=\hat{\mbtheta}} = -\nabla^2 \log p(\mbtheta) - \sum_{i=1}^N \sum_{k=1}^K \nabla^2 \log p(\mby_i \g T_{\mbgamma_k}(\mbx_i),\mbtheta)|_{\mbtheta=\hat{\mbtheta}}.
\end{align}

Assuming \(\mbgamma_k \sim p(\mbgamma \g \hat{\mbphi})\) (e.g., using a point estimate of \(\mbphi\)), this approximates:
\begin{align}
\approx -\nabla^2 \log p(\mbtheta) - K \sum_{i=1}^N \nabla^2 \log \E_{p(\mbgamma|\hat{\mbphi})}[p(\mby_i \g T_\mbgamma(\mbx_i),\mbtheta)]|_{\mbtheta=\hat{\mbtheta}} \approx K \cdot \Sigma_{\text{true}}^{-1}.
\end{align}

Therefore, \(\Sigma_{\text{na\"ive}} \approx \frac{1}{K} \Sigma_{\text{true}}\).
\end{proof}

\subsection{Proof of \cref{thm:emp-bayes} (Empirical Bayes Optimality via Augmented \ELBO)} \label{sec:proof_thm_emp_bayes}

\begin{proof}
The proof leverages variational inference principles. Start with the log marginal likelihood:
\begin{align}
\log p(\mathcal{D}) = \log \int \int \int p(\mathcal{D},\mbtheta,\mbphi, \mbgamma) \, d\mbtheta \, d\mbphi \, d\mbgamma.
\end{align}
Introduce the variational distribution \(q(\mbtheta,\mbphi) = q(\mbtheta) q(\mbphi)\):
\begin{align}
\log p(\mathcal{D}) = \mathbb{E}_{q(\mbtheta,\mbphi)} \left[ \log \frac{p(\mathcal{D},\mbtheta,\mbphi)}{q(\mbtheta,\mbphi)} \right] + \text{KL}(q(\mbtheta,\mbphi) \| p(\mbtheta,\mbphi \g \mathcal{D})).
\end{align}
Since the KL divergence is non-negative, we obtain the lower bound:
\begin{align}
\log p(\mathcal{D}) \geq \mathbb{E}_{q(\mbtheta,\mbphi)} \left[ \log \frac{p(\mathcal{D},\mbtheta,\mbphi)}{q(\mbtheta,\mbphi)} \right].
\end{align}
Now, factor in the augmentation variable \(\mbgamma\). The joint likelihood is:
\begin{align}
p(\mathcal{D} \g\mbtheta,\mbphi) = \mathbb{E}_{p(\mbgamma \g\mbphi)} \left[ \prod_{i=1}^N p(\mby_i \g T_\mbgamma(\mbx_i),\mbtheta) \right].
\end{align}
Applying Jensen’s inequality:
\begin{align}
\log p(\mathcal{D} \g\mbtheta,\mbphi) \geq \mathbb{E}_{p(\mbgamma \g\mbphi)} \left[ \sum_{i=1}^N \log p(\mby_i \g T_\mbgamma(\mbx_i),\mbtheta) \right].
\end{align}
Substitute into the lower bound:
\begin{align}
\log p(\mathcal{D}) \geq \mathbb{E}_{q(\mbtheta,\mbphi)} \left[ \mathbb{E}_{p(\mbgamma \g\mbphi)} \left[ \sum_{i=1}^N \log p(\mby_i \g T_\mbgamma(\mbx_i),\mbtheta) \right] + \log \frac{p(\mbtheta) p(\mbphi)}{q(\mbtheta) q(\mbphi)} \right].
\end{align}
Rewrite:
\begin{align}
= \mathbb{E}_{q(\mbtheta)} \mathbb{E}_{q(\mbphi)} \mathbb{E}_{p(\mbgamma \g\mbphi)} \left[ \sum_{i=1}^N \log p(\mby_i \g T_\mbgamma(\mbx_i),\mbtheta) \right] - \text{KL}(q(\mbtheta) \| p(\mbtheta)) - \text{KL}(q(\mbphi) \| p(\mbphi)).
\end{align}
Thus, we arrive at the augmented \ELBO. When \(q(\mbtheta) = p(\mbtheta \g \mathcal{D})\) and \(q(\mbphi) = p(\mbphi \g \mathcal{D})\), the bound becomes tight, confirming the result.
\end{proof}

\subsection{Proof of \cref{thm:information-gain} (Information Gain from Augmentation)} \label{sec:proof_thm_info_gain}

\begin{proof}
Under a Gaussian approximation to the posterior, the entropy is proportional to the log determinant of the covariance matrix. Using the results from Theorem~\cref{thm:posterior-shrinkage}, we have:
\begin{align}
\Delta H \propto \log\det(\Sigma_{\mathrm{no aug}}) - \log\det(\Sigma_{\mathrm{aug}}),
\end{align}
where \(\Sigma_{\mathrm{no aug}} \approx H_{\mathrm{no aug}}^{-1}\), \(\Sigma_{\mathrm{aug}} \approx H_{\mathrm{aug}}^{-1}\), and \(H_{\mathrm{aug}}\) incorporates the effect of augmentation. Thus:
\begin{align}
\Delta H \approx \frac{1}{2} \log \det (H_{\mathrm{aug}} H_{\mathrm{no aug}}^{-1}) = \frac{1}{2} \log \det (I + H_{\mathrm{no aug}}^{-1} (H_{\mathrm{aug}} - H_{\mathrm{no aug}})).
\end{align}
Approximating the effect of augmentation as an effective increase in Fisher information, we obtain the stated result.
\end{proof}

\section{A Primer on PAC-Bayes Theory}
\label{app:pac-bayes-primer}

The Probably Approximately Correct (PAC)-Bayes framework, pioneered by \citep{mcallester1999pac} and further developed by \citep{catoni2007pac} among others, provides a powerful tool for deriving generalization bounds for Bayesian-inspired learning algorithms. Unlike traditional PAC learning which often focuses on a single hypothesis, PAC-Bayes theory considers a distribution over hypotheses.

\paragraph{Core Idea}
The central idea is to bound the true risk (expected loss on unseen data) of a \emph{posterior} distribution \(Q\) over a hypothesis class \(\mathcal{H}\). This bound is typically expressed in terms of the empirical risk (average loss on the training data) under \(Q\), and a complexity term that measures how much \(Q\) deviates from a data-independent \emph{prior} distribution \(P\) over \(\mathcal{H}\). The guarantee holds with high probability (at least \(1-\delta\)) over the random draw of the training dataset.

\paragraph{Key Components}
\begin{itemize}
    \item \textbf{Hypothesis Class (\(\mathcal{H}\)):} The set of all possible models (e.g., sets of parameters \(\mbtheta\)).
    \item \textbf{Prior Distribution (\(P\)):} A distribution over \(\mathcal{H}\) chosen \emph{before} observing any training data. It reflects initial beliefs about good hypotheses.
    \item \textbf{Posterior Distribution (\(Q\)):} A distribution over \(\mathcal{H}\) that is typically learned from the training data \(\mathcal{D}\). In PAC-Bayes, \(Q\) can be any distribution, not necessarily a true Bayesian posterior.
    \item \textbf{Loss Function (\(\ell(h, z)\)):} Measures the error of a hypothesis \(h \in \mathcal{H}\) on a data point \(z = (\mbx, \mby)\).
    \item \textbf{True Risk (\(R(Q)\)):} The expected loss of a hypothesis drawn from \(Q\) on the true (unknown) data distribution: \(R(Q) = \E_{h \sim Q} [\E_{z \sim \mathcal{D}_{\text{true}}} [\ell(h, z)]]\).
    \item \textbf{Empirical Risk (\(\hat{R}(Q)\)):} The average loss of a hypothesis drawn from \(Q\) on the \(N\) training samples: \(\hat{R}(Q) = \E_{h \sim Q} [\frac{1}{N} \sum_{i=1}^N \ell(h, z_i)]\).
    \item \textbf{Kullback-Leibler (KL) Divergence (\(\KL(Q \| P)\)):} Measures the "distance" or "information gain" from the prior \(P\) to the posterior \(Q\). It serves as a complexity penalty: if \(Q\) is very different from \(P\), the penalty is high.
\end{itemize}

\paragraph{A Common Form of PAC-Bayes Bound}
A typical PAC-Bayes generalization bound (e.g., McAllester's 1999 bound or variations) states that for any \(\delta \in (0, 1)\), with probability at least \(1-\delta\) over the draw of an i.i.d. training set \(\mathcal{D}\) of size \(N\), for all posterior distributions \(Q\):
\begin{equation}
    R(Q) \leq \hat{R}(Q) + \sqrt{\frac{\KL(Q \| P) + \ln(\frac{1}{\delta}) + C}{2N}}
    \label{eq:generic-pac-bayes}
\end{equation}
where \(C\) is a constant that can depend on the range of the loss or other factors (e.g., \(\ln(2\sqrt{N})\) as used in our paper, which is a common variant for empirical Bernstein bounds).

\paragraph{Interpretation and Significance}
\begin{itemize}
    \item The bound guarantees that the true risk is unlikely to be much larger than the empirical risk, plus a term that penalizes the complexity of \(Q\) relative to \(P\).
    \item It highlights a trade-off: to achieve good generalization, a learning algorithm should find a posterior \(Q\) that both fits the training data well (low \(\hat{R}(Q)\)) and does not deviate too much from the prior (low \(\KL(Q \| P)\)).
    \item The bounds are often tighter than uniform convergence bounds for complex hypothesis classes like neural networks, especially when a good prior is available.
    \item They provide a theoretical justification for regularization techniques and can guide the design of learning algorithms.
\end{itemize}

\paragraph{Relevance to This Paper}
In our work (\cref{sec:generalization}), we adapt this framework to derive generalization bounds for our augmented likelihood approach. Here, the "hypothesis" space effectively includes both the model parameters \(\mbtheta\) and the augmentation (hyper)parameters \(\mbphi\). The priors \(p(\mbtheta)\) and \(p(\mbphi)\) and the variational posteriors \(q(\mbtheta)\) and \(q(\mbphi)\) play the roles of \(P\) and \(Q\). Our Theorem~\cref{thm:pac-bayes} provides such a bound, and Theorem~\cref{thm:generalization-advantage} uses PAC-Bayes reasoning to show the theoretical advantage of our marginalized approach over na\"ive data replication. The KL terms in our augmented \ELBO (\cref{eq:aug_elbo_def}) naturally appear as complexity measures in these PAC-Bayes bounds.

\section{Algorithm and Implementation}
\label{sec:algorithm}

We now present a practical algorithm for implementing our Bayesian-optimized data augmentation approach. The algorithm employs stochastic gradient-based optimization of both model parameters and augmentation distribution parameters.

\begin{algorithm}[ht]
\caption{Augmented Variational Inference with Learned Augmentation}
\label{alg:augmented-vi}
\begin{algorithmic}[1]
\State \textbf{Input:} Dataset \(\mathcal{D} = \{(\mbx_i, \mby_i)\}_{i=1}^N\), transformation family \(T_\mbgamma(\cdot)\)
\State \textbf{Initialize:} Variational distributions \(q(\mbtheta)\) and \(q(\mbphi)\)
\While{not converged}
    \State Sample a minibatch \(\{(\mbx_i, \mby_i)\}_{i=1}^B\) from \(\mathcal{D}\)
    \State Sample model parameters \(\mbtheta \sim q(\mbtheta)\) (or use reparameterization)
    \State Sample augmentation parameters \(\mbphi \sim q(\mbphi)\) (or use reparameterization)
    \For{each \((\mbx_i, \mby_i)\) in the minibatch}
        \State Sample augmentation parameters \(\mbgamma_i \sim p(\mbgamma|\mbphi)\)
        \State Apply transformation \(\mbx_i' = T_{\mbgamma_i}(\mbx_i)\)
        \State Compute log-likelihood \(\log p(\mby_i|\mbx_i',\mbtheta)\)
    \EndFor
    \State Estimate \ELBO:
    \[
    \widehat{\mathrm{ELBO}}_{\text{aug}} = \frac{N}{B} \sum_{i=1}^B \log p(\mby_i|\mbx_i',\mbtheta) - \KL(q(\mbtheta)||p(\mbtheta)) - \KL(q(\mbphi)||p(\mbphi))
    \]
    \State Update variational parameters in \(q(\mbtheta)\) using gradient of \(\widehat{\mathrm{ELBO}}_{\text{aug}}\)
    \State Update variational parameters in \(q(\mbphi)\) using gradient of \(\widehat{\mathrm{ELBO}}_{\text{aug}}\)
\EndWhile
\State \textbf{Output:} Optimized variational distributions \(q(\mbtheta)\) and \(q(\mbphi)\)
\end{algorithmic}
\end{algorithm}

\begin{algorithm}[ht]
\caption{Partial Variational Inference with Learned Augmentation}
\label{alg:pvi-aug}
\begin{algorithmic}[1]
\State \textbf{Input:} Dataset \(\mathcal{D} = \{(\mbx_i, \mby_i)\}_{i=1}^N\), transformation family \(T_\mbgamma(\cdot)\)
\State \textbf{Initialize:} Model parameters \(\mbtheta\) and distribution \(q(\mbphi)\)
\While{not converged}
    \State Sample a minibatch \(\{(\mbx_i, \mby_i)\}_{i=1}^B\) from \(\mathcal{D}\)
    \State Sample augmentation parameters \(\mbphi \sim q(\mbphi)\)
    \For{each \((\mbx_i, \mby_i)\) in the minibatch}
        \State Sample \(\mbgamma_i \sim p(\mbgamma|\mbphi)\)
        \State Apply transformation \(\mbx_i' = T_{\mbgamma_i}(\mbx_i)\)
        \State Compute log-likelihood \(\log p(\mby_i|\mbx_i',\mbtheta)\)
    \EndFor
    \State Estimate \ELBO:
    \[
    \widehat{\mathrm{ELBO}}_{\text{aug}} = \frac{N}{B} \sum_{i=1}^B \log p(\mby_i|\mbx_i',\mbtheta) - \KL(q(\mbphi)||p(\mbphi))
    \]
    \State Update \(\mbtheta\) using gradient of \(\widehat{\mathrm{ELBO}}_{\text{aug}}\)
    \State Update \(q(\mbphi)\) using gradient of \(\widehat{\mathrm{ELBO}}_{\text{aug}}\)
\EndWhile
\State \textbf{Output:} Optimized parameters \(\mbtheta\) and distribution \(q(\mbphi)\)
\end{algorithmic}
\end{algorithm}

\subsection{Parameterization of Augmentation Distribution}

For continuous transformation parameters, we typically use a Gaussian distribution for \(p(\mbgamma \g\mbphi)\):
\begin{equation}
p(\mbgamma|\mbphi) = \mathcal{N}(\mbgamma|\mu_\mbphi, \Sigma_\mbphi),
\end{equation}
where \(\mbphi = (\mu_\mbphi, \Sigma_\mbphi)\). For \(q(\mbphi)\), we might use a Gaussian:
\begin{equation}
q(\mbphi) = \mathcal{N}(\mbphi \g \mu_q, \Sigma_q),
\end{equation}
learning \(\mu_q\) and \(\Sigma_q\). This allows for reparameterization during sampling:
\begin{equation}
\mbphi = \mu_q + \Sigma_q^{1/2} \epsilon, \quad \epsilon \sim \mathcal{N}(0, I),
\end{equation}
followed by:
\begin{equation}
\mbgamma = \mu_\mbphi + \Sigma_\mbphi^{1/2} \epsilon', \quad \epsilon' \sim \mathcal{N}(0, I).
\end{equation}

For discrete transformations, we can use a categorical distribution:
\begin{equation}
p(\mbgamma|\mbphi) = \text{Cat}(\mbgamma|\pi_\mbphi),
\end{equation}
where \(\pi_\mbphi\) represents the probabilities, and use the Gumbel-Softmax trick~\citep{jang2017categorical} for differentiable sampling.

\subsection{Practical Considerations}

\paragraph{Adaptive Variance Scheduling.} Based on \cref{cor:optimal-variance}, we can implement an adaptive schedule for the augmentation variance within \(q(\mbphi)\), adjusting the variance of \(\mbphi\) over training to balance exploration and bound tightness.

\paragraph{Marginalization Advantage Monitoring.} Following \cref{cor:marginalization-advantage}, we can monitor the marginalization advantage term \(D_\mbphi(x_i, y_i)\) during training to assess the benefit of \ourmethod over na\"ive augmentation.

\paragraph{Curvature-Aware Augmentation.} Inspired by \cref{cor:optimal-covariance}, we can adapt the augmentation distribution based on the model's sensitivity to different transformations, allocating more variance to directions where the model is approximately invariant.

\paragraph{Computational Efficiency.} For large models, we use Monte Carlo estimates with a small number of samples (e.g., one per data point per iteration) to approximate the expectations in the \ELBO. The reparameterization trick ensures low-variance gradient estimates.

\section{Additional Experimental Details for \cref{subsec:exp-imagenet}}
% \label{app:exp-details}
\label{app:imagenet:implementdetails}

We use a ResNet-50 architecture with a Bayesian linear layer at the end (for non-Bayesian case, we just use ResNet-50 without any replacements). We apply standard preprocessing for \imagenet. % \maurizio{Merge the next text with previous one}
We use the Adam optimizer with a learning rate of \num{1e-5} for model parameters. In \ourmethod, we are learning a parameter in Beta distribution for Mixup, in uniform distribution for Cutmix, and Dirichlet, depth and Beta distribution parameters jointly for Augmix augmentations. For these augmentations, we use lognormal distribution as a prior because of the simplicity. The augmentation parameters have a separate learning rate (\num{1e-3}) to facilitate faster exploration. 
We regularize the augmentation parameters with a KL weight of $\beta_{\mathrm{kl\_aug}}=1$, balancing data-fit and prior
alignment. For all methods, we include a small KL weight $\beta_{\mathrm{kl\_net}}=10^{-4}$ on the model parameters to maintain
a Bayesian prior but it works with any weight on the model parameters % (we can  increase in case of giving more uncertainty for models). 
This acts as a Bayesian regularizer on the final layer weights, preventing overfitting within that layer and ensuring consistency with the variational Bayesian framework \citep{blundell2015weight}. % Added justification for model KL weight
Training proceeds for \num{30} epochs with a batch size of \num{256}.

\paragraph{Evaluation Metrics.} We compute the Expected Calibration Error (ECE) by dividing predictions into 10 bins based on confidence and measuring the difference between average confidence and accuracy in each bin:
\begin{align}
\text{ECE} = \sum_{i=1}^{10} \frac{|B_i|}{n} |\text{acc}(B_i) - \text{conf}(B_i)|,
\end{align}
where \(B_i\) is the set of examples in bin \(i\), \(n\) is the total number of examples, \(\text{acc}(B_i)\) is the accuracy in bin \(i\), and \(\text{conf}(B_i)\) is the average confidence in bin \(i\).

For out-of-distribution detection, we use the AUROC metric, which measures the area under the ROC curve when using predictive entropy as the detection score:
\begin{align}
H[p(y|x)] = -\sum_{c=1}^C p(\mby =c \g \mbx) \log p(\mby =c \g \mbx).
\end{align}

Higher entropy indicates higher uncertainty, which should correlate with out-of-distribution examples.

\section{Additional Results on Different Types of Data Augmentation}
\label{app:additional-results}

\subsection{Learning Geometric Augmentation for \cifar Classification}
\label{sec:exp_cifar10}

\begin{figure}[ht]
    \centering
    \includegraphics[width=0.95\textwidth]{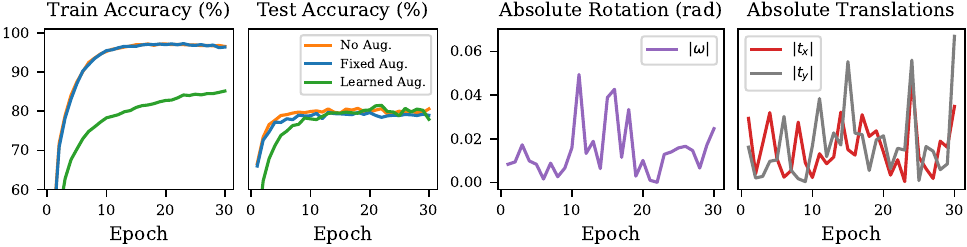}
%    %\vspace{-1ex}
    \caption{\emph{(Left Two)} Convergences of training and test accuracy on \cifar. \ourmethod generelizes better than the other approaches.  (\emph{Right Two}) Evolutions of the data augmentation parameters. }
    \label{fig:cifar10_accuracy}
    
   \vspace{-3.4ex}
\end{figure}

%\vspace{-0.9ex}
%\vspace{-1.7ex}

\begin{wraptable}{r}{0.35\textwidth}
% \vspace{-6.5mm}
    \centering
    \caption{\cifar classification. 
    % \ECE is computed with $100$ \MC samples. 
    Fixed~Aug uses fixed rotation $\omega=0.1$ and translations of $0.1$.
    }
    \label{tab:cifar10_results}
    \scalebox{.9}{
    \begin{tabular}{@{}lcc}
    \toprule
    \textbf{Method} & \textbf{Acc (\%)} & \textbf{ECE} ($\downarrow$)  \\
    \midrule
    No Aug  & \(80.90\) & \(0.092\) \\
    Fixed Aug & \(80.73\) & \(0.088\) \\
    \ourmethod& \(\textbf{81.35}\) & \(\textbf{0.017}\)  \\
    \bottomrule
    \end{tabular}
    }
\end{wraptable}

\paragraph{Setups.}
We use a ResNet-18 architecture with a Bayesian linear layer at the end. We apply standard preprocessing: normalization with mean (0.4914, 0.4822, 0.4465) and standard deviation (0.2023, 0.1994, 0.2010). % \maurizio{Merge the next text with previous one}
We use the Adam optimizer with a learning rate of \num{1e-4} for model parameters. In \ourmethod, we are learning $\mbgamma = \{\omega, t_x, t_y \}$ jointly, where $\omega$ is rotation (radians), and $t_x$ and $t_y$ are horizontal and vertical shifts. The augmentation parameters have a separate learning rate (\num{1e-2}) to facilitate faster exploration. 
We regularize the augmentation parameters with a KL weight of $\beta_{\mathrm{kl\_aug}}=1$, balancing data-fit and prior
alignment. For all methods, we include a small KL weight $\beta_{\mathrm{kl\_net}}=0.1$ on the model parameters to maintain
a Bayesian prior but it works with any weight on the model parameters. This acts as a Bayesian regularizer on the final layer weights, preventing overfitting within that layer and ensuring consistency with the variational Bayesian framework \citep{blundell2015weight}. % Added justification for model KL weight
Training proceeds for \num{30} epochs with a batch size of \num{128}.

% \begin{enumerate}
%     \item \textbf{No Augmentation (No Aug):} The model is trained on the original dataset without any geometric transformations.
    % \item \textbf{Fixed Geometric Augmentation (Fixed Aug):} We fix a rotation parameter $\mbtheta = 0.1$ radians
    % and horizontal/vertical translations of $0.1$ (in normalized coordinates), applying this affine transformation
    % identically to each training example.
    % \item \textbf{Learnable Geometric Augmentation (Learned Aug):} We treat the rotation and translations as
    % latent variables (cf.\ \cref{sec:method}), each with a Gaussian variational distribution that is regularized by
    % a $\mathrm{KL}$ term (relative to a prior). This allows the network to learn an optimal distribution
    % over geometric transformations that best explains the data.
% \end{enumerate}
% In the case of {\bf Fixed Aug} we set the rotation parameter to $0.1$ radians and horizontal/vertical translations parameters to $0.1$. 

\paragraph{Results.}
We assess calibration using $100$ \MC samples. %, averaging the softmax outputs across samples.
\cref{fig:cifar10_reliability} presents reliability diagrams for \textbf{No Aug}, \textbf{Fixed Aug}, and \ourmethod, revealing that the learned augmentation strategy yields
the lowest calibration error (ECE), with the reliability curve closely aligning with perfect calibration.
\cref{tab:cifar10_results} summarizes the final \ECE values, confirming that \ourmethod leads to more
accurate confidence estimates than fixed or no augmentation.
Moreover, \cref{fig:cifar10_accuracy} (second panel) shows test accuracy over time: the learned augmentation generalizes better, while \textbf{No Aug} and \textbf{Fixed Aug} exhibit overfitting and poorer generalization. 
% In \cref{fig:cifar10_accuracy} we also report the evolution of the learned rotation (the third panel) and translation (the fourth panel)
% parameters, demonstrating that the augmentation distribution broadens during training to better capture variations
% in the data.
% %
% For completeness, we also report a comparison between \ourmethod and Bayesian optimization in \cref{app:additional-results}, showing that \ourmethod achieves superior performance at a lower computational cost.
% \cref{fig:cifar10_accuracy} (third and fourth panels) shows the learned rotation and translation parameters evolving, indicating the augmentation distribution broadens during training to better capture data variation. Additionally, \cref{app:additional-results} compares \ourmethod to Bayesian optimization, demonstrating superior performance with lower computational cost. 

\subsection{Exploring different intensities of Gaussian translations}
\label{app:cifar10:svhn}

We use the same implementation details as in \cref{sec:exp_cifar10}, except that we choose Gaussian Translation as an augmentation parameter and validate on different values of $K$ in na\"ive augmentation, and different prior variance $\sigma$ in \ourmethod. As an OOD data, we choose the \textsc{svhn} dataset, since this mismatch makes it a widely adopted benchmark for OOD testing of classifiers trained on \cifar.

\begin{table}[ht]
\centering
\caption{Effect of marginalization vs. na\"ive augmentation with different numbers of augmentations per example on \cifar using Pretrained Bayesian ResNet-18 (last layer) and Gaussian Translation after 30 epochs. Test accuracy and \ECE are on \cifar, and OOD AUROC is on the \textsc{svhn} dataset. }
\label{tab:ablation-marginalization}
\scalebox{.82}{
    \begin{tabular}{lcccc}
    \toprule
    Method & Test Acc (\%) & \ECE $\downarrow$ & OOD AUROC\\
    \midrule
    No Aug & 94.09 & 0.0381 & 0.9069\\
    na\"ive Aug (K=2) & 95.03 & 0.0298 & 0.9425\\
    na\"ive Aug (K=5)  & 95.21 & 0.0327 & 0.9383\\
    na\"ive Aug (K=10) & 93.75 & 0.0424 & 0.9560\\
    \ourmethod ($\sigma$ = 0.1)     & 93.30  & \textbf{0.0192} & 0.9446 \\
    \ourmethod ($\sigma$ = 0.5)     & 93.87  & \textbf{0.0165} & \textbf{0.9576} \\
    \ourmethod ($\sigma$ = 1)     & 90.25  & \textbf{0.0175} & \textbf{0.9647} \\
    \bottomrule
    \end{tabular}
}
\end{table}

\paragraph{Results.} \cref{tab:ablation-marginalization} shows that \ourmethod allows us to get much better calibration than na\"ive and no augmentation cases. Because of the overcounting problem in na\"ive case, it obviously consumes around K times more time than our approach demostrating that we can get good generalization and better robustness for OOD in short time with our approach.

% \section{Broader Impacts}

% \ourmethod has significant potential for improving the reliability of Bayesian deep learning in high-stakes applications, such as medical imaging, autonomous driving, and scientific discovery, and in any safety-critical applications involving decision-making.
% %
% The ability to learn optimal augmentation strategies from data also reduces the need for manual tuning, making Bayesian methods more accessible to practitioners across domains.

%%%%%%%%%%%%%%%%%%%%%%%%%%%%%%%%%%%%%%%%%%%%%%%%%%%%%%%%%%%%

\section{Additional Experiment on \imagenet using Resnet-50}

Here we use OPTIMA for Imagenet in order to learn Mixup, Cutmix and Augmix augmentations.

\subsection{Implementation Details for \ourmethod with Mixup}
\label{app:impldetails:learnable_mixup}

We evaluate our \ourmethod framework with the Learnable Mixup augmentation on the \imagenet \citep{deng2009imagenet} dataset for image classification. Performance is assessed on the standard \imagenet validation set, and robustness is measured on the \imagenetc \citep{hendrycks2019benchmarking} benchmark.

\paragraph{Model Architecture and Preprocessing.}
We employ a ResNet-50 architecture \citep{he2016deep}, initialized with pretrained weights from \texttt{torchvision.models.ResNet50\_Weights.IMAGENET1K\_V2}. The final fully connected layer is replaced with a new linear layer mapping to the 1000 \imagenet classes. 
For input preprocessing during training, images are transformed by a \texttt{RandomResizedCrop} to \(224 \times 224\) pixels followed by a \texttt{RandomHorizontalFlip}. 
Validation and test images are resized to 256 pixels on their shorter edge and then center-cropped to \(224 \times 224\). 
All images are subsequently converted to tensors and normalized using the standard \imagenet mean \(\boldsymbol{\mu}_{\text{ImageNet}} = (0.485, 0.456, 0.406)\) and standard deviation \(\boldsymbol{\sigma}_{\text{ImageNet}} = (0.229, 0.224, 0.225)\).

\paragraph{Learnable Mixup Augmenter.}
In the \ourmethod Mixup variant, the Mixup hyperparameter \(\alpha\) (controlling the Beta distribution \(\text{Beta}(\alpha, \alpha)\) from which the mixing coefficient \(\lambda\) is sampled) is made learnable. We parameterize a Normal distribution over \(\text{logit}(\alpha)\) with learnable mean \(\mu_{\ell\alpha}\) and learnable log standard deviation \(\log \sigma_{\ell\alpha}\), where \(\ell\alpha = \text{logit}(\alpha)\).
The initial value for \(\mu_{\ell\alpha}\) is set to \(\text{logit}(0.2)\), corresponding to an initial \(\alpha_{\text{init}} = 0.2\). The initial \(\log \sigma_{\ell\alpha}\) is set to \(\log(0.1)\), promoting a small initial variance for the learned distribution over \(\text{logit}(\alpha)\).
A prior distribution \(p(\text{logit}(\alpha))\) is defined as \(\mathcal{N}(\text{logit}(\alpha_{\text{init}}), \sigma_p^2)\), where the prior standard deviation \(\sigma_p = 2.0\). 
The KL divergence between the learned variational posterior \(q(\text{logit}(\alpha) | \mu_{\ell\alpha}, \sigma_{\ell\alpha}^2)\) and this prior is added to the training objective, weighted by the hyperparameter \texttt{beta\_augmenter\_reg}. Sampled \(\lambda\) values are clamped to the range \([10^{-6}, 1 - 10^{-6}]\) for numerical stability.

\paragraph{Training Configuration.}
Models were trained for 10 epochs\footnote{While longer training (e.g., 90-100 epochs) is standard for \imagenet, these experiments were conducted for 10 epochs to demonstrate the behavior of the learnable augmentation parameters and compare against fixed augmentation under identical short-run conditions.} using the AdamW optimizer.
The base learning rate for the ResNet-50 parameters was set to \num{1e-4}. The learnable parameters of the Mixup augmenter (\(\mu_{\ell\alpha}, \log \sigma_{\ell\alpha}\)) utilized a learning rate of \num{1e-3} (10 times the base learning rate). A cosine annealing learning rate scheduler with warm restarts (\texttt{CosineAnnealingWarmRestarts}) was employed, with parameters `$T_0=10$` epochs, `$T_{mult}=2$`, and `$eta_{min}$` set to \(\nicefrac{1}{100}\) of the initial learning rate.
The weight decay for network parameters (\texttt{beta\_network\_reg}) was \(0.01\). The coefficient for the KL divergence term of the augmentation parameters (\texttt{beta\_augmenter\_reg}) was \(1.0\).
Training was performed with a global batch size of \(256\) distributed across 4 NVIDIA A100 GPUs using Distributed Data Parallel (DDP). We used a precision of `"16"` (interpreted as 16-bit native mixed precision) and set `$torch.set\_float32\_matmul\_precision('medium')$`. Gradient clipping was applied with a maximum norm of \(1.0\). The number of data loader workers was set to \(8\) per GPU process.

\paragraph{Baselines.}
We compare \ourmethod Mixup against:
\begin{itemize}[noitemsep,topsep=0pt,leftmargin=*]
    \item \textbf{Fixed Mixup}: Standard Mixup augmentation with a fixed \(\alpha = 0.2\). The training setup (optimizer, scheduler, epochs, batch size) was identical to that of \ourmethod Mixup, excluding elements specific to learnable augmentation parameters.
    \item \textbf{Pretrained ResNet-50 (No Augmentation Eval)}: The ResNet-50 model with weights from \texttt{torchvision.models.ResNet50\_Weights.IMAGENET1K\_V2}, evaluated directly on the validation and \imagenetc sets without any fine-tuning under our experimental setup. This serves as a standard reference.
\end{itemize}

\subsection{Implementation Details for \ourmethod with CutMix}
\label{app:impldetails:learnable_cutmix}

For evaluating \ourmethod with CutMix, we follow a similar experimental setup on the \imagenet \citep{deng2009imagenet} dataset, with robustness assessed on \imagenetc \citep{hendrycks2019benchmarking}.

\paragraph{Model Architecture and Preprocessing.}
We use the ResNet-50 architecture \citep{he2016deep} pretrained with \texttt{torchvision.models.ResNet50\_Weights.IMAGENET1K\_V2}, replacing the final classifier layer for the 1000 \imagenet classes. Input preprocessing during training includes \texttt{RandomResizedCrop} to \(224 \times 224\) and \texttt{RandomHorizontalFlip}. Validation and test images are resized (256 shorter edge) and center-cropped to \(224 \times 224\). Standard \imagenet normalization is applied.

\paragraph{Learnable CutMix Augmenter.}
In CutMix \citep{yun2019cutmix}, a patch from one image is pasted onto another, and labels are mixed proportionally to the area of the patches. The mixing ratio \(\lambda\) (determining the area of the first image to keep, and thus (1-\(\lambda\)) is the area of the patch from the second image) is typically sampled from a \(\text{Beta}(\alpha, \alpha)\) distribution.
For our \ourmethod CutMix, this \(\alpha\) parameter of the Beta distribution is made learnable. We parameterize a Normal distribution over \(\log(\alpha)\) with learnable mean \(\mu_{\log\alpha}\) and learnable log standard deviation \(\log \sigma_{\log\alpha}\).
The initial value for \(\mu_{\log\alpha}\) is set to \(\log(1.0)\), corresponding to an initial \(\alpha_{\text{init}} = 1.0\) (a common default for CutMix). The initial \(\log \sigma_{\log\alpha}\) is set to \(\log(0.1)\).
A prior distribution \(p(\log(\alpha))\) is defined as \(\mathcal{N}(\log(\alpha_{\text{init}}), \sigma_p^2)\), with prior standard deviation \(\sigma_p = 2.0\).
The KL divergence between the learned variational posterior for \(\log(\alpha)\) and this prior is added to the training loss, weighted by \texttt{beta\_augmenter\_reg}. The sampled \(\alpha\) values are clamped to \([10^{-4}, 100.0]\) before being used in the Beta distribution. The resulting mixing coefficient \(\lambda_{\text{final}}\) (coefficient for the first image's label) is determined by the actual area of the pasted patch after clipping to image boundaries.

\paragraph{Training Configuration.}
Models were trained for \(N_{\text{epochs}}\) epochs (e.g., 15) using the AdamW optimizer. The base learning rate for network parameters was \num{1e-4}, while the learnable CutMix parameters (\(\mu_{\log\alpha}, \log \sigma_{\log\alpha}\)) used a learning rate of \num{1e-3}. A \texttt{CosineAnnealingWarmRestarts} learning rate scheduler was used (`T\_0=10` or `15`, `T\_{mult}=2`, `eta\_{min}`=\(\nicefrac{1}{100}\) of initial LR).
Network weight decay (\texttt{beta\_network\_reg}) was \(0.01\), and the KL coefficient (\texttt{beta\_augmenter\_reg}) was \(1.0\).
Training used a global batch size of \(256\) on 4 NVIDIA A100 GPUs with DDP, `"16"` precision, `$torch.set\_float32\_matmul\_precision('medium')$`, and gradient clipping at \(1.0\). Data loader workers were set to \(8\) per GPU.

%---

\subsection{Implementation Details for \ourmethod with AugMix (Learnable Severity + JSD)}
\label{app:impldetails:learnable_augmix_jsd}

We evaluate \ourmethod by learning a component of the AugMix \citep{hendrycks2019augmix} augmentation strategy, specifically its overall \texttt{aug\_severity}, while also employing the Jensen-Shannon Divergence (JSD) consistency loss. Experiments are conducted on \imagenet \citep{deng2009imagenet} and \imagenetc \citep{hendrycks2019benchmarking}.

\paragraph{Model Architecture and Preprocessing.}
The model is a ResNet-50 \citep{he2016deep} initialized with \texttt{torchvision.models.ResNet50\_Weights.IMAGENET1K\_V2}, with the final classifier layer adapted for 1000 classes. 
During training, input PIL images undergo \texttt{RandomResizedCrop} to \(224 \times 224\) and \texttt{RandomHorizontalFlip}. These PIL images are then passed to our \texttt{LearnableAugMixSeverityJSDAugmenter} module.
Validation and test images use standard resizing, center cropping, and \texttt{ToTensor} conversion, followed by \imagenet normalization.

\paragraph{Learnable AugMix Severity + JSD Augmenter.}
The \texttt{LearnableAugMixSeverityJSDAugmenter} is implemented as an \texttt{nn.Module}.
\begin{itemize}[noitemsep,topsep=0pt,leftmargin=*]
    \item \textbf{AugMix Core}: For each input PIL image, three views are generated: the original, and two independently augmented versions using AugMix. Each AugMix version is a convex combination (\(m \sim \text{Beta}(1,1)\)) of the original image and a mixture of \(K=\texttt{augmix\_mixture\_width}\) (default 3) augmentation chains. Each chain consists of \(D\) (default random 1-3, controlled by \texttt{augmix\_mixture\_depth}) basic operations (e.g., rotate, shear, color jitter) sampled randomly. Mixing weights for chains \(w_k\) are from \(\text{Dirichlet}(\mathbf{1})\). All PIL operations and the final conversion to tensors (for each of the three views) happen within this augmenter module, ensuring output tensors are on the correct device. We utilize a predefined list of tensor-based augmentation operations where possible to improve performance over PIL-only operations.
    \item \textbf{Learnable Severity}: The overall intensity of the basic augmentations, \texttt{aug\_severity} (typically a value between 0-10), is made learnable. We parameterize a Normal distribution over \(\log(\text{aug\_severity})\) with learnable mean \(\mu_{\log S}\) and learnable log standard deviation \(\log \sigma_{\log S}\). The initial \(\mu_{\log S}\) corresponds to an \texttt{initial\_aug\_severity} of \(3.0\), and initial \(\log \sigma_{\log S} = \log(0.1)\). The prior for \(\log(\text{aug\_severity})\) is \(\mathcal{N}(\log(\text{initial\_aug\_severity}), \sigma_{pS}^2)\) with \(\sigma_{pS} = \texttt{prior\_severity\_std\_learnable\_aug}\) (default 1.0). A KL divergence term, weighted by \texttt{beta\_augmenter\_reg}, regularizes these learned severity parameters. The sampled severity is clamped to \([0.1, 10.0]\).
    \item \textbf{JSD Loss}: The three output image tensors (original, AugMix view 1, AugMix view 2) are passed through the network. A JSD consistency loss is calculated between their softmax predictions, weighted by \texttt{beta\_jsd} (default 12.0), and added to the primary cross-entropy loss (calculated on the original view).
\end{itemize}
The data loader for training uses a custom collate function to provide a list of PIL images to the augmenter.

\paragraph{Training Configuration.}
Training was conducted for \(N_{\text{epochs}}\) epochs (in our case 6 epochs because of the computational complexity related to tensor and PIL tranformations) with the AdamW optimizer. The base learning rate for network parameters was \num{1e-4}. The learnable severity parameters (\(\mu_{\log S}, \log \sigma_{\log S}\)) used a learning rate of \num{1e-3}. A \texttt{CosineAnnealingWarmRestarts} scheduler was used (`T\_{0}=10` or `15`, `T\_{mult}=2`, `eta\_{min}`=\(\nicefrac{1}{100}\) of initial LR).
Network weight decay (\texttt{beta\_network\_reg}) was \(0.01\). The KL coefficient for severity parameters (\texttt{beta\_augmenter\_reg}) was \(1.0\). The JSD loss coefficient (\texttt{beta\_jsd}) was \(12.0\).
Training used a global batch size of \(128\) (reduced due to processing three views) on 4 NVIDIA A100 GPUs with DDP, `16` precision, `$torch.set\_float32\_matmul\_precision('medium')$`, and gradient clipping at \(1.0\). Data loader workers were \(8\) per GPU.

\subsection{Evaluation and Software/Hardware for all these methods.}

Models are evaluated on the standard \imagenet validation set for top-1 accuracy and cross-entropy loss. Robustness is assessed on the \imagenetc benchmark, reporting the normalized mean Corruption Error (mCE normalized by AlexNet baseline) across all corruptions and severities. For \imagenetc, images are processed using the same validation transforms as for the clean \imagenet validation set. All final reported evaluations are performed on a single GPU using 32-bit floating-point precision.

Experiments were conducted using PyTorch version 2.0.1 and PyTorch Lightning version 2.1.0. Training and evaluation four utilized NVIDIA A100 (80GB) GPUs.

%The code to reproduce all the results code can be found in \url{https://github.com/imadik31/OPTIMA}.

\subsection{Experimental Results}
\label{sec:app:resnet50-nonbayesian}

\begin{table}[ht]
\centering
\caption{The result of \imagenet and \imagenet-C using ResNet-50 for each augmentations. We evaluate the average test error for each corruption type}
\label{tab:cifar10:bo}
\scalebox{.99}{
    \begin{adjustbox}{width=\textwidth}
    \begin{tabular}{lccccccccccccccccc}
    \toprule
    Method & Test Acc (\%) & mCE (normalized) (\%) & gaussian & shot & impulse & defocus & glass & motion & zoom & snow & frost & fog & brightness & contrast & elastic & pixelate & jpeg\\
    \midrule
    No Aug & 60.8 & 76.7 & 71 &	73 & 76 & 61 & 73 &	61 &	64 & 67 & 62 & 54 &	32 & 61 & 55 & 55 & 47 \\
    
    Mixup \citep{zhang2017mixup} & 77.9 & - & - & - & - & - & - & - & - & - & - & - & - & - & - & - & - \\
    
    \ourmethod Mixup (10 epochs) & \textbf{79.31} & 68.41 & 55 & 57 & 59 & 62 & 73 & 59 & 59 & 62 & 51 & 44 & 31 & 48 & 55 & 50 & 43 \\

    Cutmix \citep{yun2019cutmix} & 78.6 & - & - & - & - & - & - & - & - & - & - & - & - & - & - & - & - \\

    \ourmethod Cutmix (15 epochs) & \textbf{79.62} & 70.6 & 68 & 68.7 & 68.8 & 74.6 & 88.6 & 75.7 & 73.4 & 73.2 & 70.8 & 59 & 54.1 & 61.3 & 83.6 & 70.5 & 69.2 \\
    
    Augmix \citep{hendrycks2019augmix} & 77.53 & 65.3 & - & - & - & - & - & - & - & - & - & - & - & - & - & - & - \\

    \ourmethod Augmix (6 epochs) & \textbf{78.19} & 68.21 & 57.37 & 58.26 & 60.79 &	60.98 &	72.04 &	56.77 &	54.54 &	59.76 &	54.58 &	44.72 &	30.13 &	45.37 &	54.75 &	51.66 &	44.71\\

    \bottomrule
    \end{tabular}
    \end{adjustbox}
}
\end{table}

\paragraph{Results.} In \cref{tab:cifar10:bo}, we can see that \ourmethod allows us to get better test accuracy on clean data with non-Bayesian ResNet-50. \ourmethod Mixup, Cutmix and Augmix are beating the baseline results within (5-15 epochs). The mCE of \ourmethod Augmix is lower than the benchmark. 
This can be explained by the very few training epochs (6 epochs) which we could run due to the computational complexity of this experiment; 
%since we are training \imagenet by transferring images into PIL image therefore it leads to some difficulties with working on CPU and GPU - 
with our computational resources it takes around 15 hours for one full epoch.

\section{Additional Experimental Details for \cref{subsec:sst5}. Token-dropout implementation details}

\label{app:sst5}

\paragraph{Parameterization.}
The augmentation module applies token dropout with probability
\[
p_{\mathrm{drop}} = p_{\max}\,\sigma(s),
\]
where $s$ is a trainable scalar (initialized at $s_0 = -2$ in our
implementation) and $\sigma(\cdot)$ is the logistic function. The constant
$p_{\max}$ sets an upper bound on the amount of dropout; we use $p_{\max}=0.5$.

\paragraph{Prior.}
We place a Gaussian prior directly on $p_{\mathrm{drop}}$,
\[
p(p_{\mathrm{drop}}) \propto
\exp\!\left(-\tfrac{1}{2}\tfrac{(p_{\mathrm{drop}} - \mu)^2}{\sigma^2}\right),
\]
implemented as a quadratic penalty in the ELBO objective. We use
$\mu \in \{0.1, 0.3\}$ to represent weak and strong prior preferences for
token dropout, and $\sigma = 0.1$.

\paragraph{OPTIMA initialization.}
All methods begin from the same initial dropout rate
$p_{\mathrm{drop}} = p_{\max}\,\sigma(s_0) \approx 0.04$, ensuring a mild
initial augmentation.

\paragraph{Baselines.}
We compare the following:
(i) \emph{No Aug} ($p_{\mathrm{drop}} = 0$);
(ii) \emph{Fixed Aug}, using the same initialization as OPTIMA;
(iii) \emph{Fixed Aug (Matched)}, using OPTIMA’s learned dropout;
(iv) \emph{BO-Fixed}, selecting $p_{\mathrm{drop}}$ via validation NLL over
a grid in $[0,0.3]$.

\paragraph{Optimizers.}
DistilBERT is trained with learning rate $2 \times 10^{-5}$, and $s$ is
trained with learning rate $5\times 10^{-2}$. Further hyperparameters are
unchanged from standard HuggingFace defaults.

\section{Broader Impact}
% \paragraph{Broader impact.}

% \bahien{I would move this section to Appendix}

\ourmethod has significant potential for improving the reliability of Bayesian deep learning in high-stakes applications, such as medical imaging, autonomous driving, and scientific discovery. %, and in any safety-critical applications involving decision-making.
The ability to learn optimal augmentation strategies from data also reduces the need for manual tuning, making Bayesian methods more accessible to practitioners across domains.

\section{Reproducibility Statement}
All experiments are fully described in the submission, including dataset details, hyperparameters, and training procedures. The accompanying code is provided to ensure that our results can be independently reproduced.

\section{The Use of Large Language Models (LLMs)}
We used large language models (LLMs) solely for non-substantive assistance, including grammar refinement and summarizing relevant literature. All research ideas, analyses, and conclusions are the authors' own.

\end{document}